\documentclass{article}

\usepackage[preprint]{corl_2026} 

\usepackage[numbers]{natbib}
\usepackage{multicol}
\usepackage{bbding}
\usepackage{multirow}
\usepackage{color}
\usepackage[most]{tcolorbox}
\usepackage{graphicx}
\usepackage{xcolor}
\usepackage{pifont}
\usepackage{dsfont}
\usepackage{colortbl, booktabs}
\usepackage{comment}
\usepackage{cuted}
\usepackage{caption}
\usepackage{subcaption}
\usepackage{booktabs}
\usepackage{array} 
\usepackage{amssymb}
\usepackage{tabularx}
\usepackage[normalem]{ulem}
\usepackage{amsmath}
\usepackage{wrapfig}    
\usepackage{adjustbox}
\usepackage{url}
\usepackage{array}
\usepackage{caption}

\title{KAI: A Kinematic-Aware Interface for Data-Efficient Articulated Object Manipulation}

\author{
  Yaping Li$^{1,2}$ \quad
  Zhaxizhuoma$^{2,3}$ \quad
  Qiaojun Yu$^{2,*}$ \quad
  Jia Zeng$^{2,*}$ \quad
  Dahua Lin$^{1,2,*}$ \quad
  Jiangmiao Pang$^{2,*}$ 
  \vspace{1mm} \\
  $^{1}$The Chinese University of Hong Kong \quad
  $^{2}$Shanghai AI Laboratory \quad
  $^{3}$Shanghai Jiao Tong University
  \vspace{1mm} \\
  $^{*}$Corresponding authors
  \vspace{1mm} \\
  Project page: \url{https://li-yaping.github.io/KAI/}
}

\begin{document}
\maketitle


\begin{figure*}[h]
  \centering
  \vspace{-4mm}
  \includegraphics[width=1.0\linewidth]{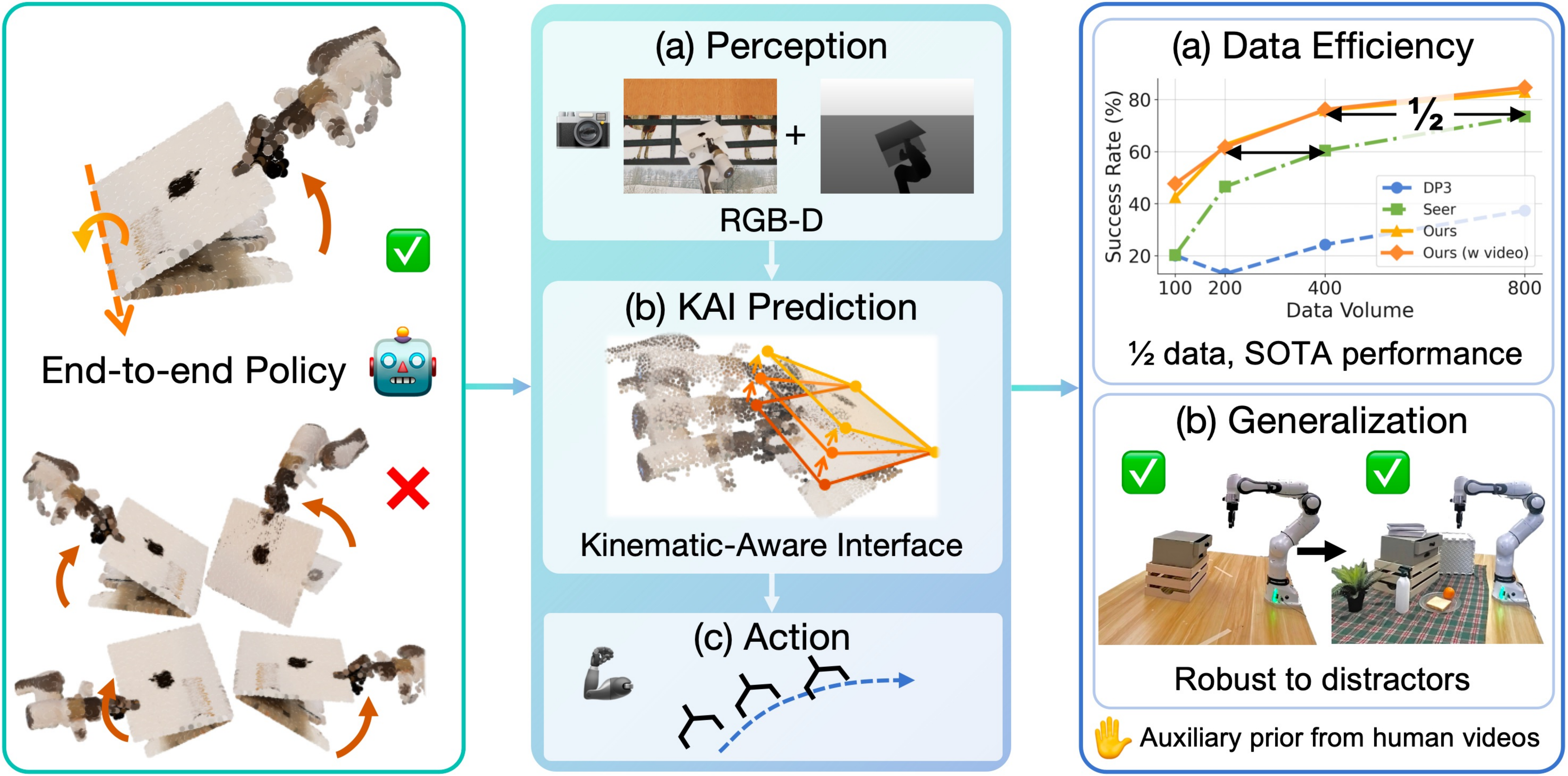}
  \captionof{figure}{\textbf{Overview.} (a) End-to-end policies map observations to actions without modeling kinematic structure, relying on large demonstration sets. (b) KAI injects an explicit structured representation that encodes articulated object kinematics into policy learning. (c) This design yields strong sample efficiency and robust generalization under diverse distractions.}
  \label{fig:teaser}
\end{figure*}


\begin{abstract}
Articulated object manipulation requires an understanding of kinematic structure that is difficult and costly to learn from robot demonstrations alone. We introduce the Kinematic-Aware Articulation Interface (KAI), a structured intermediate representation that captures the kinematic structure of articulated objects. By embedding interpretable geometric and kinematic priors into policy learning, KAI provides a strong inductive bias aligned with the underlying structure of articulated motion. This design effectively improves sample efficiency, with gains particularly pronounced in low-data regimes: across six simulation tasks, our method achieves an average success rate of 82.9\%, matching or surpassing baseline performance while using only half the demonstration data. Our method also exhibits robust generalization to unseen backgrounds and visual distractors, transferring from a single clean training environment to cluttered real-world scenes. KAI's action-agnostic design further enables co-training with human interaction videos to enhance real-world robustness: under diverse visual distractions, our method with video co-training achieves over 70\% average success rate.
\end{abstract}



\section{Introduction}
\vspace{-1mm}

Articulated objects---doors, drawers, laptops, microwaves---are ubiquitous in human environments. Despite their diversity in appearance and function, these objects share a fundamental property: their motion is governed by kinematic constraints. A door rotates about a fixed hinge; a drawer translates along a constrained rail. Reliable manipulation therefore requires understanding what part of the object moves and in what direction. Acquiring such structural understanding from limited interaction data, however, remains an open challenge for robot learning.

End-to-end imitation learning methods learn direct perception-to-action mappings without explicitly modeling kinematic structure, forcing the policy to infer joint mechanics implicitly from data. Under limited demonstrations, this often leads to reliance on superficial visual cues and brittle generalization when backgrounds or object appearances change. To improve data efficiency, prior works have introduced intermediate representations such as affordance maps \cite{ning2023where2explore,geng2023rlafford,wu2021vat,wang2022adaafford,mo2021where2act}, point flow \cite{eisner2022flowbot3d,zhang2023flowbot++,xu2022universal}, and grasp pose \cite{cui2025gapartmanip,morlans2024grasp,yu2024gamma}. While these representations constrain the learning space, they typically describe where to interact or what motion should result, rather than directly encoding the underlying kinematic structure of the object. We suggest that a more effective inductive bias for articulated object manipulation is one that captures the kinematic constraints of articulated joints---an interface that explicitly represents kinematic structure itself.

To address these limits, we introduce \textbf{Kinematic-Aware Articulation Interface (KAI)}, a structured intermediate representation that encodes the kinematic structure of an articulated object by localizing interaction keypoints on the moving part and predicting future displacement trajectories. This explicit geometric and kinematic prior enables the policy to learn more fundamental, generalizable representations of articulated motion from substantially fewer demonstrations. Across six simulation tasks, our method achieves an average success rate of 82.9\%, matching or surpassing baseline performance while using only half the demonstration data. The gains are particularly pronounced in low-data regimes. Our method also exhibits robust generalization to unseen backgrounds and visual distractors, transferring from a single clean training environment to cluttered real-world scenes. KAI's action-agnostic design further enables co-training with human interaction videos to enhance real-world robustness: under diverse visual distractions, our method with video co-training achieves over 70\% average success rate.

Our contributions are summarized as follows:
\vspace{-1mm}
\begin{itemize}
    \item We propose the \textbf{Kinematic-Aware Articulation Interface (KAI)}, a structured intermediate representation that explicitly encodes the kinematic structure of articulated objects. Across six simulation tasks, our method achieves an average success rate of 82.9\%.
    \item KAI achieves notable gains in \textbf{sample efficiency}: our method matches or surpasses baseline performance while using only half the demonstration data, and the gains are particularly pronounced in low-data regimes.
    \item KAI enables \textbf{robust generalization} from single-scene training to diverse unseen backgrounds and distractors. Its action-agnostic design further allows co-training with human interaction videos, yielding over 70\% average success rate in cluttered real-world scenes.
\end{itemize}

\vspace{-1mm}

\section{Related Work}
\label{sec:related_work}

\vspace{-1mm}

\subsection{Articulated Object Manipulation}
\vspace{-1mm}

Manipulating articulated objects requires reasoning about their kinematic structure, which has been approached from two broad directions. Prediction-Planning Methods explicitly estimate interface outputs---such as joint parameters---from observations and then plan actions accordingly~\cite{jain2021screwnet,jain2022distributional,zeng2021visual,li2020category,yu2024gamma,geng2023gapartnet,wang2024rpmart,eisner2022flowbot3d,zhang2023flowbot++,cui2025gapartmanip}. End-to-End Learning-Based Methods train models to map raw observations directly to actions~\cite{mo2021where2act,wu2021vat,xu2022universal,wang2025articubot,geng2023rlafford}. While end-to-end methods offer flexibility and have demonstrated strong generalization potential, they typically do not model the object's kinematic structure explicitly. This places a heavy burden on data collection: without an inductive bias that captures the underlying kinematic structure, the model requires a large volume of costly robot demonstrations to learn a robust policy. This motivates incorporating an explicit kinematic intermediate representation into the end-to-end pipeline.

\begin{figure*}[t]
  \centering
  \includegraphics[width=1.0\linewidth]{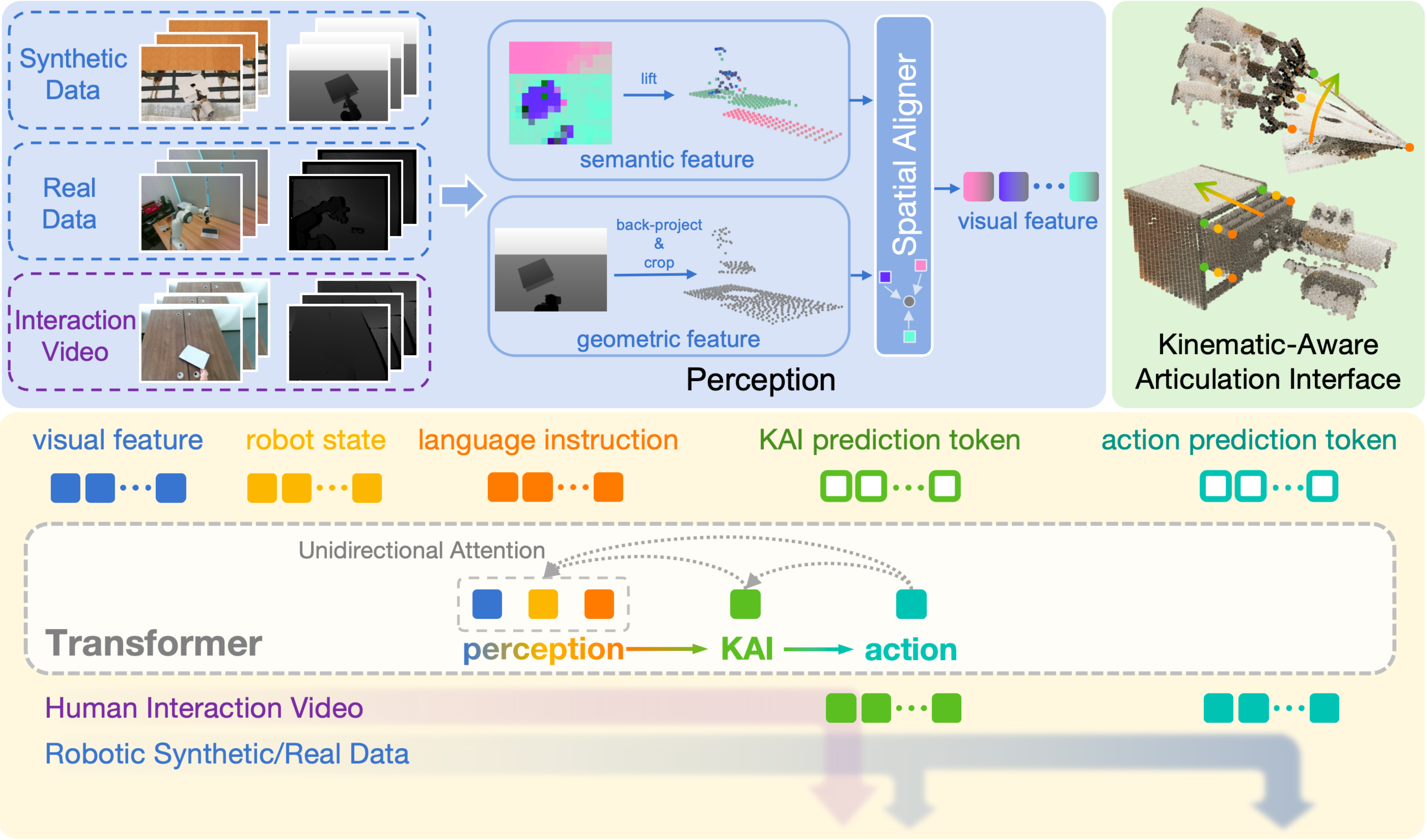}
   \caption{\textbf{Method overview.} (a) \textbf{Perception:} RGB and point cloud inputs are encoded and fused into visual features. (b) \textbf{KAI:} The KAI module predicts keypoint locations and motion displacements to represent the kinematic structure of the articulated object. (c) \textbf{Action:} The action decoder generates robot actions conditioned on both visual features and KAI predictions.}
   \label{fig:method}
   \vspace{-4mm}
\end{figure*}

\vspace{-1mm}

\subsection{Representations for Data-Efficient Learning}

\vspace{-2mm}

To improve data efficiency, a significant research direction introduces structured intermediate representations that decompose the manipulation problem by predicting explicit task-relevant properties from observations. Prior works have explored various forms of representations, such as affordance maps \cite{ning2023where2explore,geng2023rlafford,wu2021vat,wang2022adaafford,mo2021where2act}, point flow \cite{eisner2022flowbot3d,zhang2023flowbot++,xu2022universal}, grasp pose \cite{cui2025gapartmanip,morlans2024grasp,yu2024gamma}, and direct joint parameter estimation \cite{yu2024gamma,wang2024articulated,yu2025uniaff}. These representations help structure the learning space to some extent, but they primarily describe where to interact or what motion should result, rather than directly encoding the kinematic structure that governs how the object moves. This motivates a representation that explicitly captures the underlying kinematic structure of articulated objects.

\vspace{-1mm}

\section{Method}
\label{sec:method}

\vspace{-2mm}

In this section, we present the implementation details of our method. First, we introduce the design of KAI (Section~\ref{sec:Kinematic-Aware Articulation Interface}). Next, we describe the overall model architecture (Section~\ref{sec:Model Architecture}). Finally, we explain the co-training strategy that enables learning from human interaction videos (Section~\ref{sec:Co-Training Strategy}).

\vspace{-1mm}

\subsection{Kinematic-Aware Articulation Interface}
\label{sec:Kinematic-Aware Articulation Interface}

\vspace{-1mm}

\begin{wrapfigure}{r}{0.5\textwidth}
    \vspace{-10mm}  
    \centering
    \begin{subfigure}{0.49\linewidth}
        \centering
        \includegraphics[width=\linewidth]{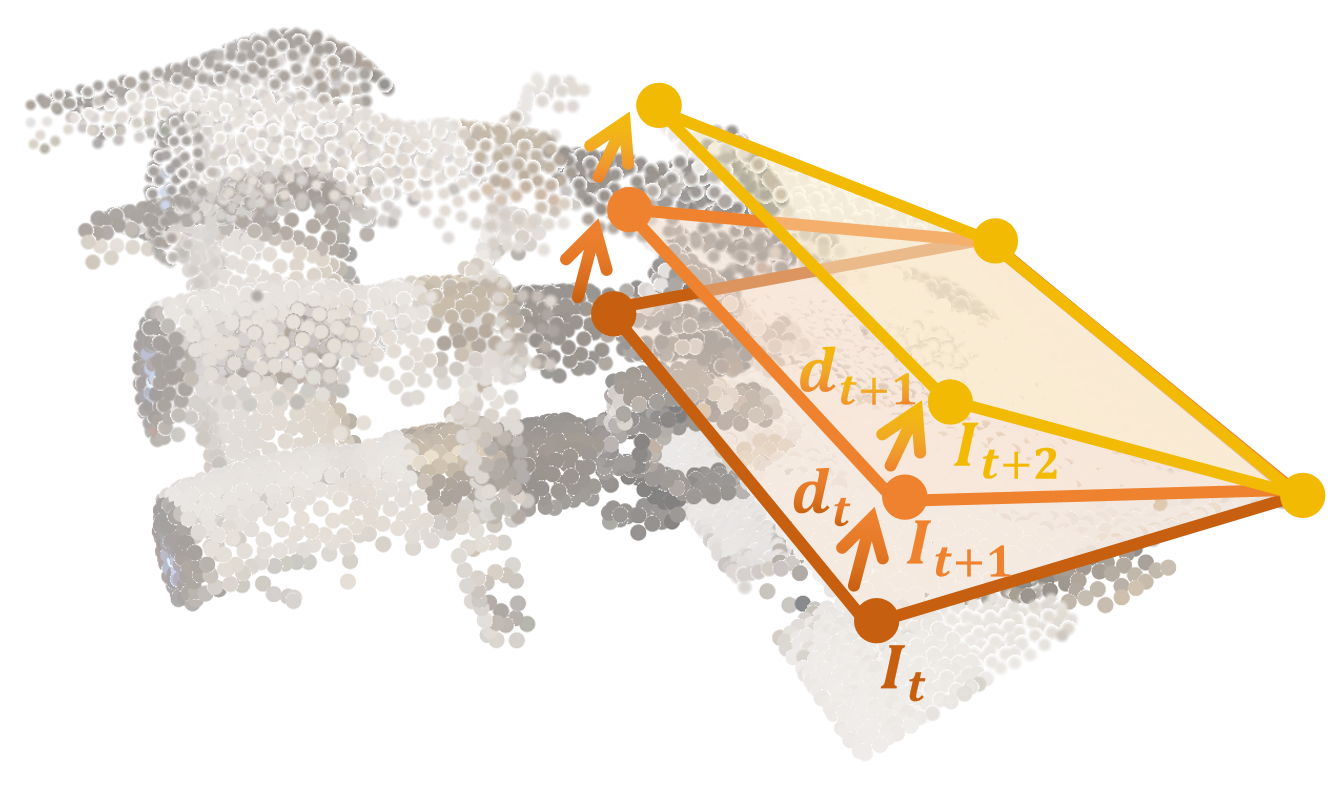}
    \end{subfigure}
    \begin{subfigure}{0.49\linewidth}
        \centering
        \includegraphics[width=\linewidth]{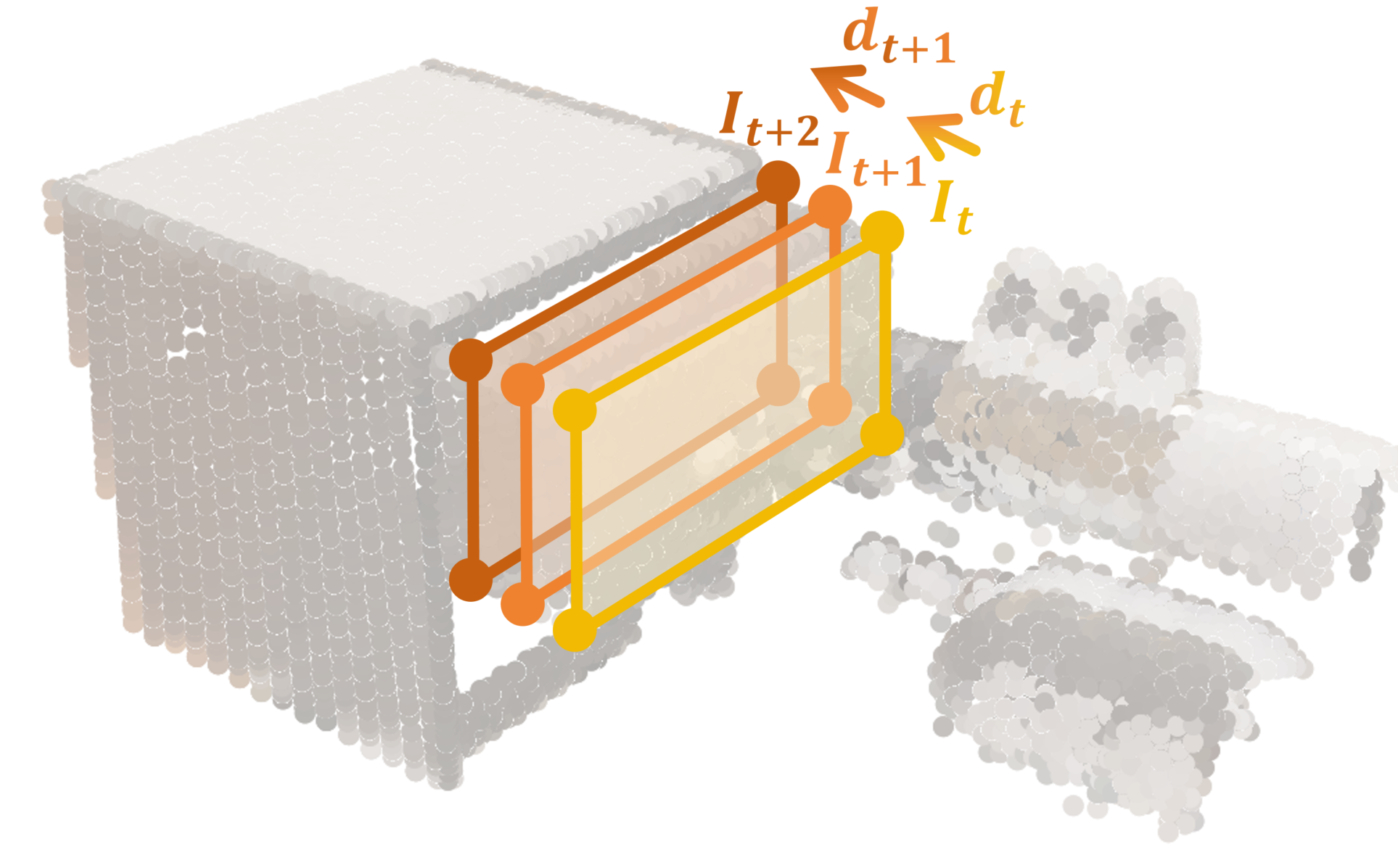}
    \end{subfigure}
    \vspace{-6mm}
    \caption{\textbf{Instantiation of KAI.} The KAI representation visualized on two manipulation tasks (laptop opening and drawer closing).}
    \label{fig:interface_diagram}
    \vspace{-5mm}
\end{wrapfigure}

KAI represents the kinematic structure of an articulated object through $K$ interaction keypoints on the moving part. For each keypoint $i \in \{1,\dots,K\}$, given an observation at timestep $t$, KAI predicts a sequence of future locations $\mathbf{l}^i_{t+1}, \dots, \mathbf{l}^i_{t+N}$ and corresponding displacements $\mathbf{d}^i_{t}, \dots, \mathbf{d}^i_{t+N-1}$ over the next $N$ timesteps, where $\mathbf{d}^i_{t+\delta} = \mathbf{l}^i_{t+\delta+1} - \mathbf{l}^i_{t+\delta}$ for $\delta \in \{0, \dots, N-1\}$. As illustrated in Figure~\ref{fig:interface_diagram} with $K=4$, this couples the part's pose with its motion pattern into a unified representation.

To ensure that predicted keypoints respect the kinematic constraints of articulated motion, we regularize the keypoint set $\mathcal{K}$ with a geometric loss $\mathcal{L}_{\text{geo}}$ that enforces structural consistency and motion coherence:
\begin{equation}
 \mathcal{L}_{\text{geo}} = \lambda_{s} \mathcal{L}_{\text{struct}} + \lambda_{m} \mathcal{L}_{\text{motion}},
\end{equation}
where $\mathcal{L}_{\text{struct}}$ enforces isometric constraints to preserve pairwise distances among keypoints over time, and $\mathcal{L}_{\text{motion}}$ regularizes displacements according to joint type.

The structural loss $\mathcal{L}_{\text{struct}}$ enforces isometric constraints to preserve the object's shape over time. By penalizing changes in the relative distances between keypoints, we prevent non-physical warping or deformation during interaction:
\begin{equation}
 \mathcal{L}_{\text{struct}} = \sum_{\delta=0}^{N-1} \sum_{i,j \in \mathcal{K}} \left| \|\mathbf{l}^i_{t+\delta+1} - \mathbf{l}^j_{t+\delta+1}\|_2 - \|\mathbf{l}^i_t - \mathbf{l}^j_t\|_2 \right|^2.
\end{equation}

For prismatic joints, the motion is constrained to pure translation. We implement a translation invariance prior by minimizing the variance of displacement vectors across the keypoint set $\mathcal{K}$, ensuring all points move coherently along the joint axis:
\begin{equation}
 \mathcal{L}_{\text{motion}}^{\text{pris}} = \sum_{\delta=0}^{N-1} \sum_{i \in \mathcal{K}} \left\| \mathbf{d}^i_{t+\delta} - \bar{\mathbf{d}}_{t+\delta} \right\|_2^2, \quad \text{where } \bar{\mathbf{d}}_{t+\delta} = \frac{1}{|\mathcal{K}|} \sum_{j \in \mathcal{K}} \mathbf{d}^j_{t+\delta}.
\end{equation}

For revolute joints, the motion is modeled as a constrained rotation. We designate a subset of keypoints $\mathcal{K}_a$ as stationary anchors located on the rotation axis, and define $\mathcal{K}_m$ as the remaining moving points. The loss penalizes any displacement of the anchor points and enforces that the orthogonal projection of each moving point onto the rotation axis remains constant over time:
\begin{equation}
\mathcal{L}_{\text{motion}}^{\text{rev}} = \sum_{\delta=0}^{N-1} \left( \sum_{a \in \mathcal{K}_a} \|\mathbf{d}^a_{t+\delta}\|_2^2 + \sum_{i \in \mathcal{K}_m} \| \operatorname{proj}(\mathbf{l}^i_{t+\delta+1}) - \operatorname{proj}(\mathbf{l}^i_{t+\delta})\|_2^2 \right),
\end{equation}
where $\operatorname{proj}(\cdot)$ denotes the orthogonal projection of a point onto the rotation axis.




\subsection{Model Architecture}
\label{sec:Model Architecture}
\vspace{-1mm}

At time step $t$, our model takes an observation $\mathbf{o}_t$ comprising RGB-D images, the robot proprioceptive state $\mathbf{s}_t$, and a natural language instruction $\mathbf{t}_t$ as input. It outputs predictions for the KAI---parameterized future keypoint locations $\{ \hat{\mathbf{l}}^i_{t+\delta+1} \}_{\delta=0}^{N-1}$ and corresponding displacements $\{ \hat{\mathbf{d}}^i_{t+\delta} \}_{\delta=0}^{N-1}$ for $i \in \{1,\dots,K\}$---and a chunk of robot actions $\{ \hat{\mathbf{a}}_{t+\delta+1} \}_{\delta=0}^{N-1}$  over $N$ timesteps.

\textbf{Perception.}
As illustrated in Figure~\ref{fig:method}, our model processes vision, text, and robot state inputs.
Inspired by RISE-2~\cite{fang2025airexo}, our perception module fuses 2D semantic features with 3D geometric features. Semantic features are extracted from the RGB image via the DINOv2~\cite{oquab2023dinov2} and lifted to 3D by projection onto the point cloud. A sparse 3D encoder processes the point cloud in parallel to capture geometric structure. The two representations are then fused by a spatial aligner into a unified set of visual tokens. Language instructions are tokenized by the CLIP~\cite{radford2021learning} text encoder, and robot proprioceptive states are embedded through a multi-layer perceptron (MLP).

\textbf{Phased-Reasoning Architecture.}
We design a phased-reasoning architecture that follows a ``Perception--KAI--Action'' workflow. KAI serves as a structured intermediate representation between perception and action. After the perception phase, KAI tokens attend exclusively to the perceptual tokens (vision, language, and robot state), extracting features for predicting keypoint locations and motion displacements without access to action-level information. The resulting KAI tokens are then combined with the original perceptual tokens and fed into the action decoder. In the action phase, the action decoder generates robot actions conditioned on both perceptual and KAI tokens. As illustrated in Figure~\ref{fig:method}, the information flow is enforced via unidirectional attention masks: KAI tokens can attend to perceptual tokens but not to action tokens. By decoupling structural understanding from control, this architecture supports learning from heterogeneous data sources, including action-agnostic human interaction videos.

\newcolumntype{Y}{>{\centering\arraybackslash}X}

\subsection{Co-Training Strategy} 
\label{sec:Co-Training Strategy}
\vspace{-1mm}

Because KAI encodes object kinematics in a form independent of specific embodiment, its predictions can be supervised directly from human interaction videos without requiring robot action labels. We use the HOI4D dataset~\cite{liu2022hoi4d} as the source of such videos. Supervision for KAI is derived from the bounding boxes of articulated parts, an annotation readily available in many interaction datasets.

Specifically, the model $\pi_{\theta}$ is trained to predict KAI from both human video and robot action data in a unified format. The KAI prediction is supervised by the loss $\mathcal{L}_{\text{KAI}}$, which comprises three terms:
\begin{equation}
  \mathcal{L}_{\text{location}} = \left\| \pi_{\theta}\left( \{ \hat{\mathbf{l}}^i_{t+\delta+1} \}_{\delta=0}^{N-1} \mid \mathbf{o}_t, \mathbf{s}_t, \mathbf{t}_t \right) - \{ \mathbf{l}^i_{t+\delta+1} \}_{\delta=0}^{N-1} \right\|^2_2,
  \label{eq:location_loss}
\end{equation}
\begin{equation}
  \mathcal{L}_{\text{motion}} = \left\| \pi_{\theta}\left( \hat{\mathcal{D}}_t \mid \mathbf{o}_t, \mathbf{s}_t, \mathbf{t}_t \right) - \mathcal{D}_t \right\|^2_2,
  \label{eq:motion_loss}
\end{equation}
\begin{equation}
  \mathcal{L}_{\text{KAI}} = \alpha_1 \mathcal{L}_{\text{location}} + \alpha_2 \mathcal{L}_{\text{motion}} + \alpha_3 \mathcal{L}_{\text{geo}},
  \label{eq:kai_loss}
\end{equation}
where $\hat{\mathcal{D}}_t = \{ \hat{\mathbf{d}}^i_{t+\delta} \}_{\delta=0}^{N-1}$ and $\mathcal{D}_t = \{ \mathbf{d}^i_{t+\delta} \}_{\delta=0}^{N-1}$ for $i \in \{1,\dots,K\}$ denote the predicted and ground-truth displacement sequences, respectively.
For samples from the action dataset, the model further predicts the future action chunk conditioned on the predicted KAI representation. The action prediction is supervised by:
\begin{equation}
  \mathcal{L}_{\text{action}} = \mathcal{L}_{\text{reg}} \left( \pi_{\theta}( \{ \hat{\mathbf{a}}_{t+\delta+1} \}_{\delta=0}^{N-1} \mid \mathbf{o}_t, \mathbf{s}_t, \mathbf{t}_t, \{ \hat{\mathbf{l}}^i_{t+\delta+1} \}_{\delta=0}^{N-1}, \hat{\mathcal{D}}_t ) , \{ \mathbf{a}_{t+\delta+1} \}_{\delta=0}^{N-1} \right),
  \label{eq:action_loss}
\end{equation}
where the action $\hat{\mathbf{a}}$ is decomposed into arm and gripper actions. The arm action is optimized using a Smooth L1 loss for continuous motion control, while the gripper action employs a Binary Cross Entropy loss for discrete command classification.

During training, each mini-batch mixes samples randomly drawn from both the interaction video dataset and the action dataset. The overall training objective is:
\begin{equation}
  \mathcal{L} = \lambda_1 \mathcal{L}_{\text{KAI}} + \lambda_2 \cdot \mathbf{1}_{\text{action}} \cdot \mathcal{L}_{\text{action}},
  \label{eq:total_loss}
\end{equation}
where the indicator $\mathbf{1}_{\text{action}}$ equals 1 for samples from the action dataset and 0 otherwise.

\vspace{-1mm}

\section{Simulation Experiments}
\label{sec:simulation_experiments}

\vspace{-1mm}

We conduct experiments to answer the following questions: (1) How does our method perform on tasks with generalization challenges? (Section~\ref{sec:Main Results and Analysis}) (2) Can the KAI improve data efficiency? (Section~\ref{sec:Data Efficiency Analysis}) (3) Do our key design choices contribute to the final performance? (Section~\ref{sec:Ablation Studies})

\vspace{-1mm}

\subsection{Experimental Setups}

\vspace{-1mm}

\textbf{Tasks and Dataset.} 
We use Isaac Sim~\cite{NVIDIA_Isaac_Sim} as the simulation platform. We implement rule-based automated data collection and evaluation for six tasks: closing and opening a laptop, a door, and a drawer, covering fundamental articulated object manipulation skills. We collect 800 expert demonstrations per task. During data collection, we randomize the object's initial state (position, orientation, and joint configuration), background scene, and lighting conditions.

\textbf{Baselines.} 
We evaluate our method against a range of state-of-the-art approaches, including the 2D image methods ACT \cite{zhao2023learning} and Seer \cite{tian2024predictive}, the 3D point cloud methods DP3 \cite{ze20243d} and RISE-2 \cite{fang2025airexo}, and ArticuBot \cite{wang2025articubot}—a 3D hierarchical approach specialized for articulated object manipulation. Detailed implementations are provided in Appendix \ref{sec:implementation_details} and \ref{sec:baseline_implementation}.

\textbf{Evaluation Metrics.}
We evaluate each method over 200 trials per task. Object initial states (position, orientation, and joint configuration) are sampled using 200 unique random seeds unseen during training, and testing is conducted across 8 unseen background scenes and lighting conditions. We use success rate as the evaluation metric, defined as the proportion of trials in which the robot completes the task within the allotted time steps.

\begin{table*}
  \caption{\textbf{Simulation experimental results.} Success rate (\%) across six tasks. ``Avg. SR'' denotes the average success rate across all tasks. The best results are bolded.}
  \label{tab:results}
  \vspace{-1mm}
  \centering
  \adjustbox{width=\textwidth}{%
    \begin{tabular}{lccccccc}
      \toprule
      Method & Avg. SR $\uparrow$ & Open Drawer & Open Door & Open Laptop & Close Drawer & Close Door & Close Laptop \\
      \midrule
      ACT~\cite{zhao2023learning} & 58.5 & 44.5 & 52.5 & 67.5 & 39.0 & 54.0 & 93.5 \\
      Seer~\cite{tian2024predictive} & 73.4 & 84.8 & \textbf{67.2} & 67.7 & 68.1 & 63.7 & 89.2 \\
      \cmidrule(r){1-8}
      DP3~\cite{ze20243d} & 37.3 & 25.0 & 12.5 & 25.5 & 2.0 & \textbf{97.0} & 62.0 \\
      RISE-2~\cite{fang2025airexo} & 22.3 & 21.0 & 9.0 & 7.5 & 21.5 & 19.0 & 56.0 \\
      ArticuBot~\cite{wang2025articubot} & 38.5 & 4.0 & 22.5 & 24.0 & 28.0 & 71.0 & 81.5 \\
      \cmidrule(r){1-8}
      \rowcolor{gray!20} Ours & 82.9 & 86.7 & 54.5 & 81.7 & \textbf{82.3} & 92.5 & \textbf{100.0} \\
      \rowcolor{gray!20} Ours (w/ video) & \textbf{84.6} & \textbf{86.8} & 66.0 & \textbf{85.0} & 81.8 & 88.8 & 99.2 \\
      \bottomrule
    \end{tabular}%
  }
  \vspace{-4mm}
\end{table*}

\vspace{-1mm}

\subsection{Simulation Results and Analysis} 
\label{sec:Main Results and Analysis}
\vspace{-3mm}
Table~\ref{tab:results} summarizes the simulation results across six tasks. Our method without video co-training achieves an average success rate of 82.9\%, already surpassing all compared baselines. Co-training with human interaction videos yields a modest further improvement to 84.6\%. The overall performance advantage validates the effectiveness of KAI as a structured intermediate representation.

The limited gain from video co-training in simulation (+1.7\%) can be attributed to the perceptual domain gap: HOI4D videos are captured with different camera properties and noise characteristics than our synthetic images. This gap narrows in real-world evaluation, where human videos and robot observations share similar visual conditions. Section~\ref{sec:real_experiments} shows that video co-training provides more substantial benefits in the real world.

\begin{wrapfigure}{R}{0.48\textwidth}
  \centering
  \vspace{-16mm}
  \includegraphics[width=\linewidth]{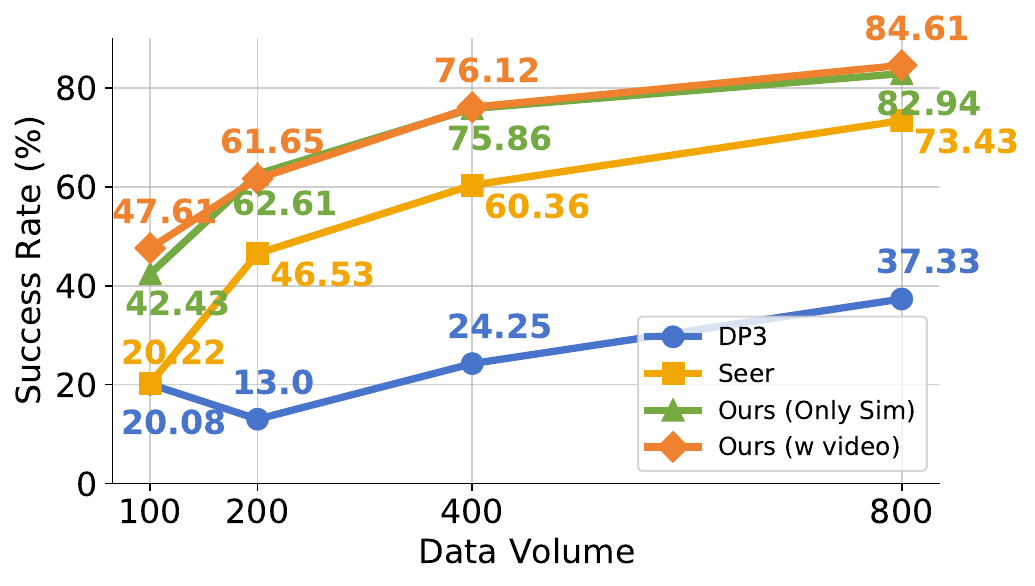}
  \vspace{-7mm}
  \caption{\textbf{Data efficiency comparison.} Success rates of our method, Seer, and DP3 across different training data volumes.}
  \label{fig:data_efficency}
  \vspace{-4mm}
\end{wrapfigure}

\vspace{-1mm}

\subsection{Data Efficiency Analysis}
\label{sec:Data Efficiency Analysis}
\vspace{-2mm}

We evaluate our method against the 2D baseline Seer~\cite{tian2024predictive} and the 3D baseline DP3~\cite{ze20243d} under varying training data sizes (100, 200, 400, and 800 demonstration episodes). As shown in Figure~\ref{fig:data_efficency}, while both our method and Seer show rapid performance improvement with increasing data before converging, DP3 exhibits much slower learning progress. Our approach consistently outperforms both baselines across all data volumes. Under the low-data setting of 100 demonstrations, our method outperforms the strongest baseline by 22 percentage points. This advantage persists at larger data scales, where our method matches or exceeds the best baseline's performance using only half the training data. These results demonstrate that KAI substantially enhances sample efficiency by embedding kinematic priors directly into the learning process, particularly in low-data regimes. Detailed results are provided in Appendix~\ref{sec:data_efficiency_details}.

\vspace{-1mm}

\subsection{Ablation Studies}
\label{sec:Ablation Studies}
\vspace{-3mm}

\begin{table*}[h]
  \caption{\textbf{Component ablation.} Success rate (\%) across six tasks for different configurations of our method. ``w/ KAI'' denotes the inclusion of KAI prediction; ``w/ $\mathcal{L}_{\text{geo}}$'' denotes the addition of geometric consistency loss; ``w/ video'' denotes co-training with human interaction videos.}
  \label{tab:ablation}
  \vspace{-1mm}
  \centering
  \resizebox{1.0\linewidth}{!}{
      \begin{tabular}{@{}ccc|ccccccc@{}}
        \toprule
        w/ KAI & w/ $\mathcal{L}_{\text{geo}}$ & w/ video & Avg. SR $\uparrow$ & Open Drawer & Open Door & Open Laptop & Close Drawer & Close Door & Close Laptop\\
        \midrule
        $\times$ & $\times$ & $\times$ & 44.1 & 16.2 & 15.5 & 40.0 & 58.2 & 50.7 & 83.8 \\
        $\checkmark$ & $\times$ & $\times$ & 75.0 & 84.5 & 37.7 & 83.7 & 80.3 & 84.5 & 79.5 \\
        $\checkmark$ & $\checkmark$ & $\times$ & 82.9 & 86.7 & 54.5 & 81.7 & \textbf{82.3} & \textbf{92.5} & \textbf{100.0} \\
        $\checkmark$ & $\checkmark$ & $\checkmark$ & \textbf{84.6} & \textbf{86.8} & \textbf{66.0} & \textbf{85.0} & 81.8 & 88.8 & 99.2 \\
        \bottomrule
      \end{tabular}
    }
    \vspace{-2mm}
\end{table*}

\textbf{Component contributions.}
Table~\ref{tab:ablation} reports ablation results isolating the contribution of each component. Removing KAI entirely yields an average success rate of 44.1\%, indicating that a vanilla end-to-end policy learns little about kinematic constraints from limited demonstrations. Adding KAI tokens alone raises the success rate to 75.0\%, the largest single improvement in the table and a clear signal that the structured kinematic representation is the primary source of gain. Incorporating the geometric consistency loss $\mathcal{L}_{\text{geo}}$ further increases performance to 82.9\%, as the physically grounded constraints regularize the predicted keypoints and their displacements. Co-training with human videos provides a modest additional improvement to 84.6\%, consistent with the perceptual domain gap discussed in Section~\ref{sec:Main Results and Analysis}. A quantitative analysis of KAI prediction accuracy and its correlation with task success is provided in Appendix~\ref{sec:kai_mde_analysis}.

\begin{wraptable}{r}{0.58\textwidth}
  \centering
  \vspace{-14pt}
  \caption{\textbf{Architectural ablation.} Success rate (\%) for alternative design choices.}
  \label{tab:ablation_supp}
  \resizebox{\linewidth}{!}{
  \begin{tabular}{lcccc}
    \toprule
    Configuration & Avg. SR $\uparrow$ & Open Drawer & Close Door & Close Laptop \\
    \midrule
    DP3 w/ DINOv2 & 17.7 & 5.0 & 33.0 & 15.0 \\
    Ours w/ Pred. Depth & 49.6 & 27.9 & 66.0 & 55.0 \\
    Ours w/o Causal Attn. & 78.5 & 89.2 & 87.5 & 58.7 \\
    \bottomrule
  \end{tabular}
  }
  \vspace{-10pt}
\end{wraptable}

\vspace{-1mm}

\textbf{Architectural design choices.}
Table~\ref{tab:ablation_supp} reports results under several alternative configurations. First, equipping DP3 with the same visual encoder used in our method (DP3 w/ DINOv2) achieves only 17.7\% average success rate. This confirms that a stronger visual encoder alone does not resolve the underlying difficulty of learning articulated motion from sparse demonstrations. Second, replacing the KAI kinematic prediction with a future depth prediction task under the same 3D perception architecture (Ours w/ Pred. Depth) yields 49.6\%, substantially below our method. Structured kinematic reasoning thus provides a more effective inductive bias than generic visual prediction for these tasks. Finally, removing the causal attention mask and allowing bidirectional attention between KAI and action tokens (Ours w/o Causal Attn.) reduces the average success rate to 78.5\%, consistent with the phased-reasoning design where kinematic inference precedes action generation. An additional evaluation under seen backgrounds and lighting is provided in Appendix~\ref{sec:seen_bg_light}.

\vspace{-3mm}

\section{Real Experiments}
\label{sec:real_experiments}

\vspace{-8mm}
\begin{figure}[h]
  \centering
  \begin{minipage}{0.46\textwidth}
    \centering
    \vspace{0.12cm}
    \includegraphics[width=\linewidth]{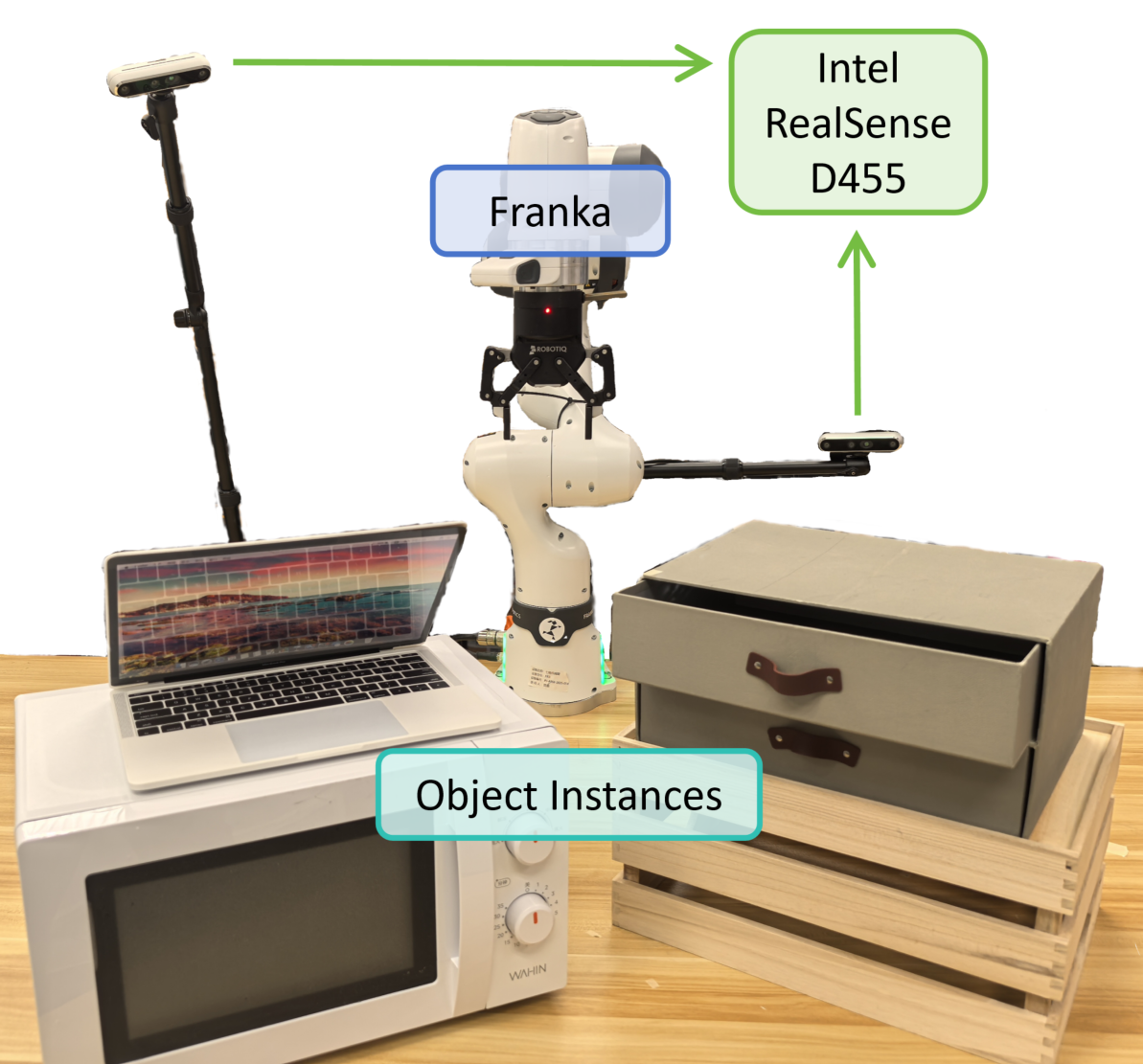}
    \vspace{-0.03cm}
    \caption{\textbf{Real-world setup.} Our setup includes a Franka Research 3 arm with a Robotiq 2F-85 gripper, complemented by two D455 cameras.}
    \label{fig:real_setup}
  \end{minipage}
  \hfill
  \begin{minipage}{0.52\textwidth}
    \centering
    \begin{subfigure}{0.96\linewidth}
      \centering
      \begin{subfigure}[b]{0.3\linewidth}
        \centering
        \includegraphics[width=\linewidth]{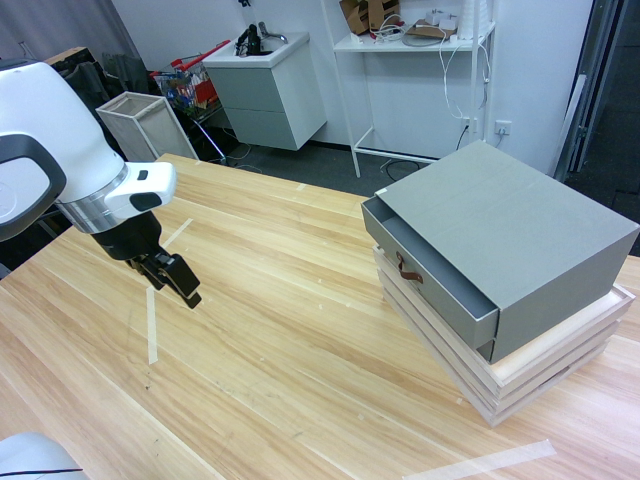}
      \end{subfigure}
      \begin{subfigure}[b]{0.3\linewidth}
        \centering
        \includegraphics[width=\linewidth]{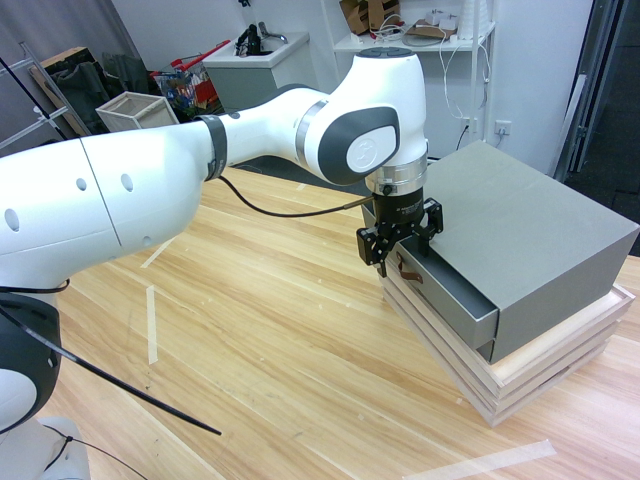}
      \end{subfigure}
      \begin{subfigure}[b]{0.3\linewidth}
        \centering
        \includegraphics[width=\linewidth]{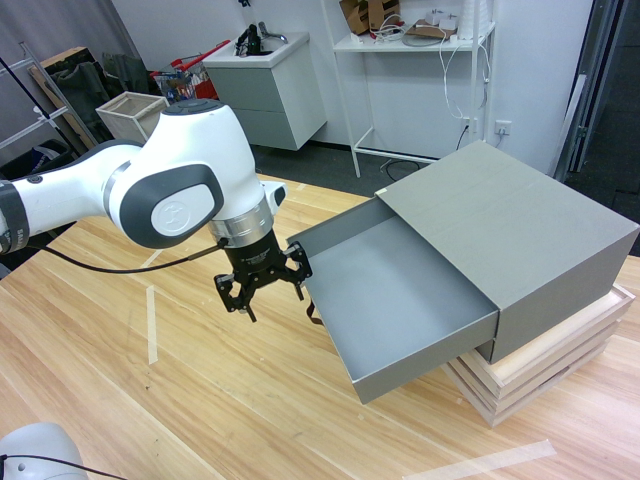}
      \end{subfigure}
      \vspace{-0.1cm}
      \caption{Opening drawer}
      \label{fig:open_drawer}
    \end{subfigure}
    
    
    \begin{subfigure}{0.96\linewidth}
      \centering
      \begin{subfigure}[b]{0.3\linewidth}
        \centering
        \includegraphics[width=\linewidth]{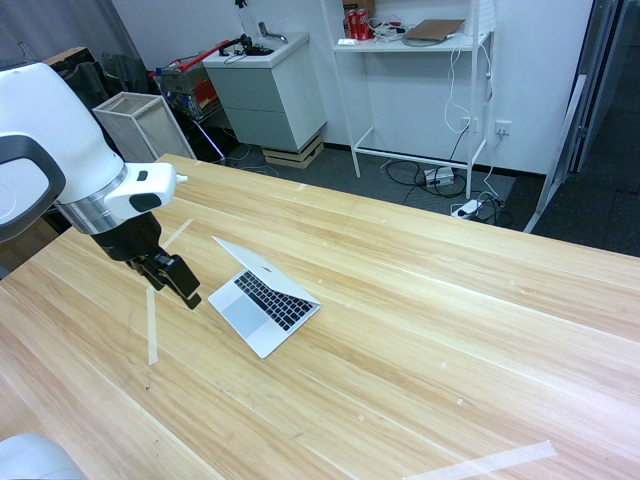}
      \end{subfigure}
      \begin{subfigure}[b]{0.3\linewidth}
        \centering
        \includegraphics[width=\linewidth]{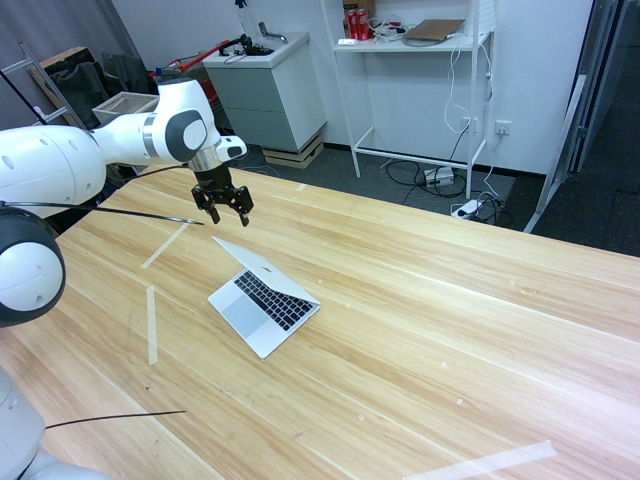}
      \end{subfigure}
      \begin{subfigure}[b]{0.3\linewidth}
        \centering
        \includegraphics[width=\linewidth]{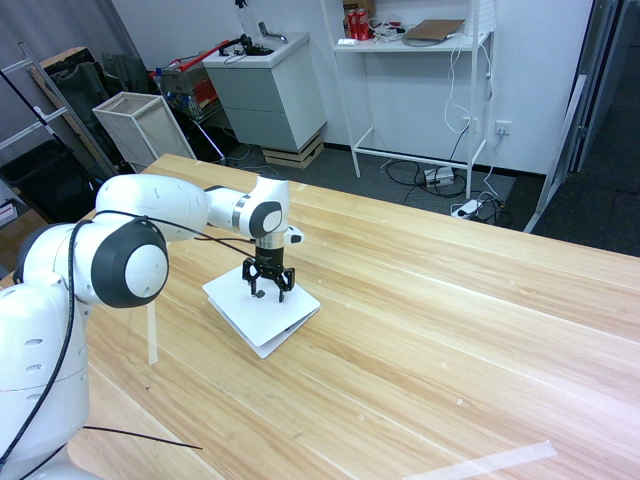}
      \end{subfigure}
      \vspace{-0.1cm}
      \caption{Closing laptop}
      \label{fig:close_laptop}
    \end{subfigure}
    
    
    \begin{subfigure}{0.96\linewidth}
      \centering
      \begin{subfigure}[b]{0.3\linewidth}
        \centering
        \includegraphics[width=\linewidth]{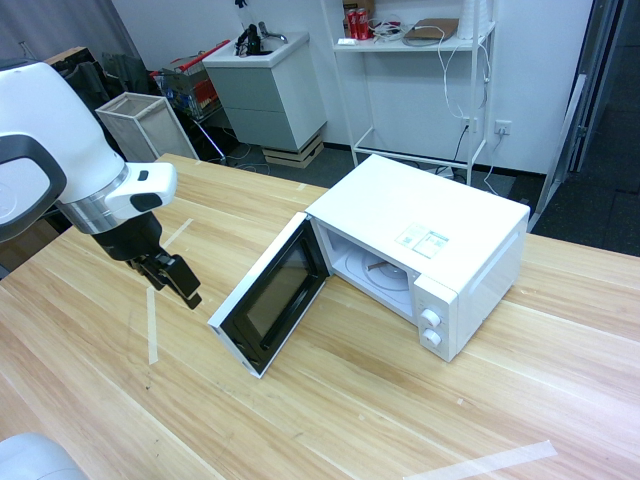}
      \end{subfigure}
      \begin{subfigure}[b]{0.3\linewidth}
        \centering
        \includegraphics[width=\linewidth]{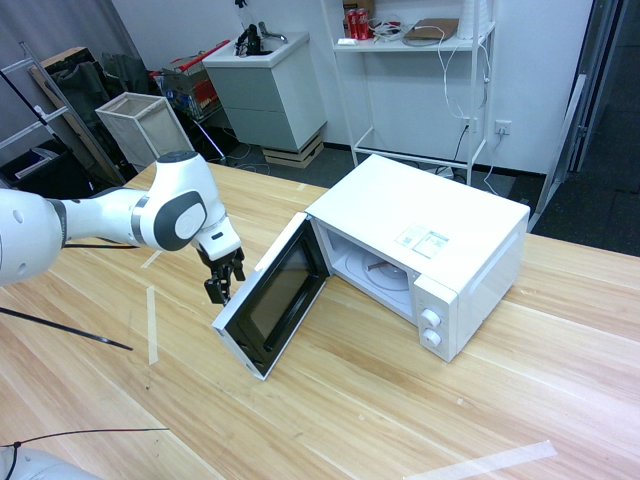}
      \end{subfigure}
      \begin{subfigure}[b]{0.3\linewidth}
        \centering
        \includegraphics[width=\linewidth]{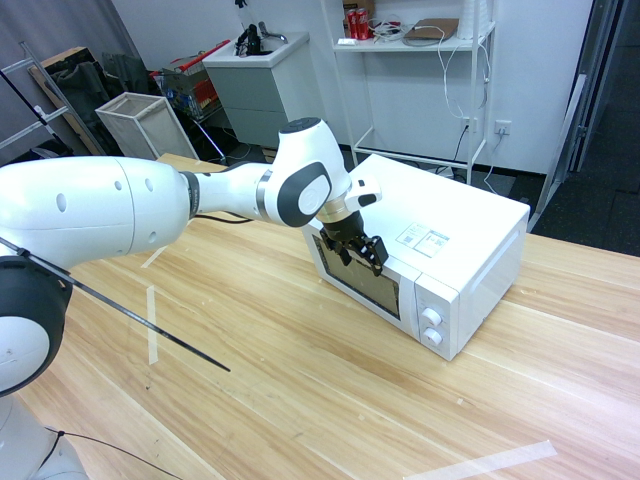}
      \end{subfigure}
      \vspace{-0.1cm}
      \caption{Closing microwave}
      \label{fig:close_microwave}
    \end{subfigure}
    \vspace{-0.12cm}
    \caption{\textbf{Sequential visualizations of three real-world tasks.} Each row shows a task progressing from left to right over time.}
    \label{fig:real_task_visualization}
  \end{minipage}
\end{figure}

\vspace{-5mm}

\subsection{Experimental Setups}
\vspace{-2mm}
\textbf{Real-world Setup.} We evaluate our method on a Franka Research 3 arm with a Robotiq 2F-85 gripper across three articulated object manipulation tasks, using two RealSense D455 cameras for RGB-D visual perception, as shown in Figure~\ref{fig:real_setup}.

\vspace{-1mm}
\textbf{Task and Dataset.} We evaluate our method on three tasks (Figure~\ref{fig:real_task_visualization}): \textit{Opening drawer}, \textit{Closing laptop} and \textit{Closing microwave}. We collect 100 demonstrations for each task in a single clean setting.

\begin{figure}[htbp]
 \vspace{-2mm}
    \centering
    \begin{subfigure}[b]{0.145\textwidth}
        \centering
        \includegraphics[width=\linewidth]{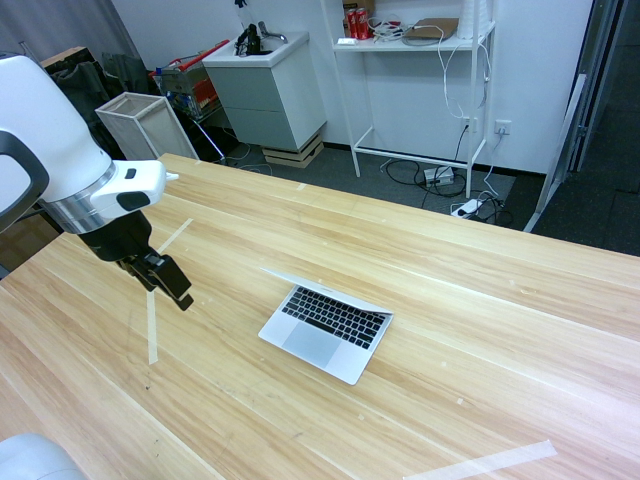}
        \caption{w/o Bg.}
        \label{fig:bg_clean}
    \end{subfigure}
    \hspace{-0.3cm}
    \begin{subfigure}[b]{0.335\textwidth}
        \centering
        \begin{minipage}[b]{0.43\textwidth}
            \centering
            \includegraphics[width=\linewidth]{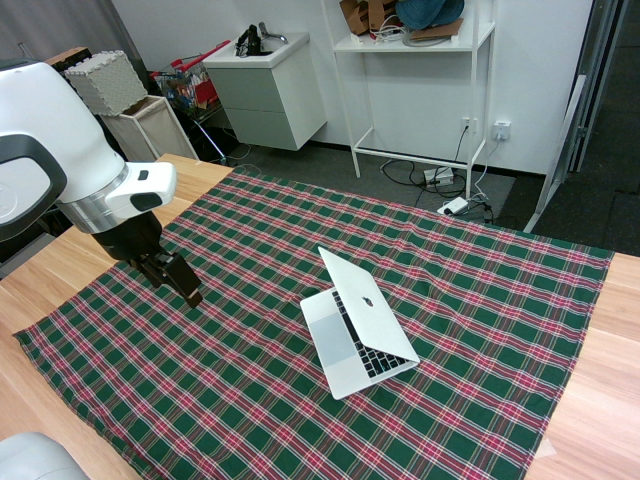}
        \end{minipage}
        \begin{minipage}[b]{0.43\textwidth}
            \centering
            \includegraphics[width=\linewidth]{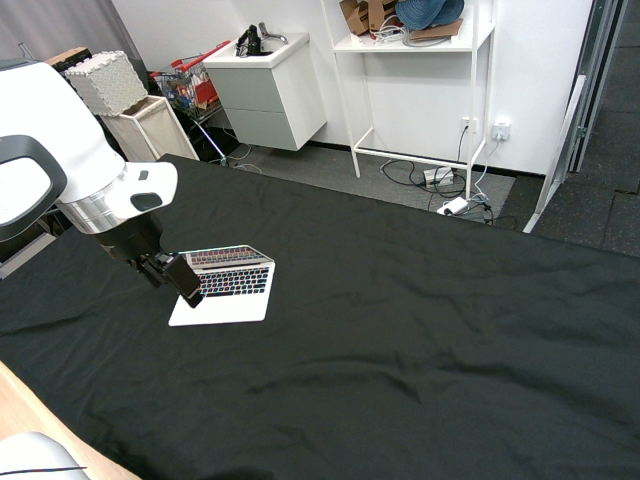}
        \end{minipage}
        \caption{w/ Background Distractions}
        \label{fig:bg_dist}
    \end{subfigure}
    \hspace{0.05cm}
    \begin{subfigure}[b]{0.145\textwidth}
        \centering
        \includegraphics[width=\linewidth]{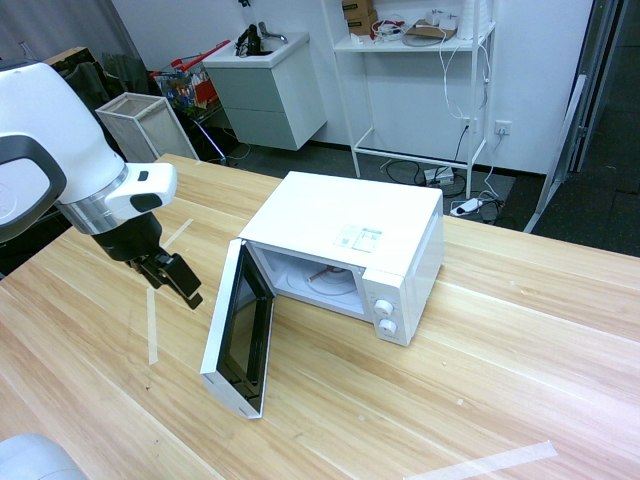}
        \caption{w/o Obj.}
        \label{fig:obj_clean}
    \end{subfigure}
    \hspace{-0.3cm}
    \begin{subfigure}[b]{0.335\textwidth}
        \centering
        \begin{minipage}[b]{0.43\textwidth}
            \centering
            \includegraphics[width=\linewidth]{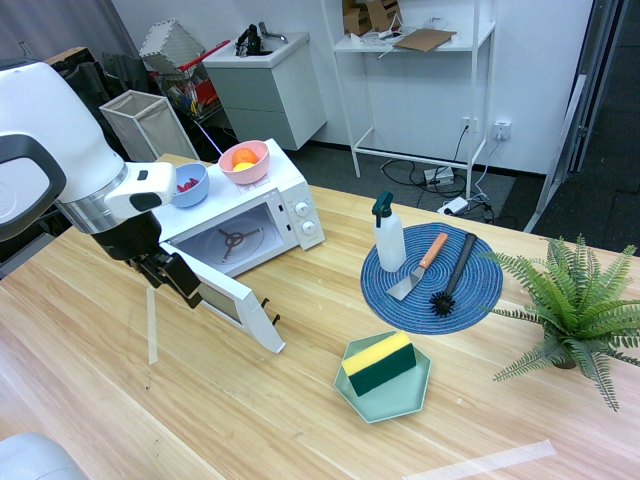}
        \end{minipage}
        \begin{minipage}[b]{0.43\textwidth}
            \centering
            \includegraphics[width=\linewidth]{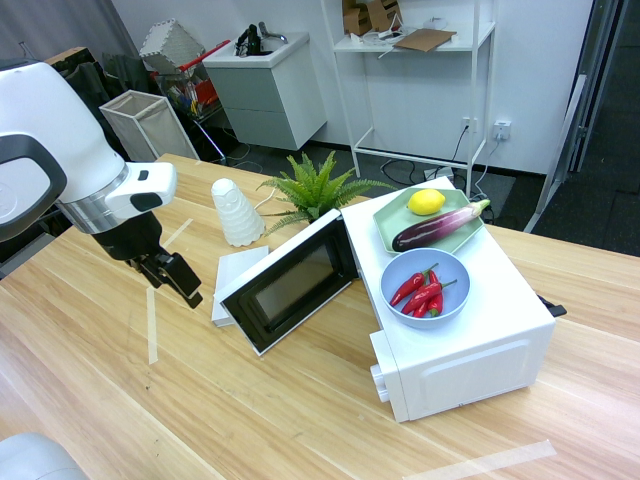}
        \end{minipage}
        \caption{w/ Object Distractors}
        \label{fig:obj_dist}
    \end{subfigure}
    \vspace{-1mm}
    \caption{Visualization of generalization tasks under different types of distractions.}
    \label{fig:distractions_combined}
    \vspace{-3mm}
\end{figure}

\vspace{-1mm}

\textbf{Evaluation Metrics.}
As illustrated in Figure~\ref{fig:distractions_combined}, we evaluate each method under three generalization settings: no distraction, background distractions, and object distractors. For each setting, five initial states are tested with three trials each. Performance is measured by two metrics: (1) Success Rate (SR), scored 100\% only upon full task completion; and (2) Localization Success Rate (Loc-SR), scored 100\% when the robot correctly reaches the object contact position. SR captures overall task execution, while Loc-SR isolates the model's ability to localize the interaction point. Per-setting initial configurations are visualized in Appendix~\ref{sec:visualizations_generalization}.

\vspace{-3mm}

\subsection{Real-world Results and Analysis}
\vspace{-6mm}

\begin{table}[htbp]
  \centering
  \caption{\textbf{Real-world results.} Success rate (SR) and localization success rate (Loc-SR) under three generalization settings. Each cell reports SR / Loc-SR (\%). The best results are bolded.}
  \vspace{1mm}
  \label{tab:real_exper}
  \resizebox{\linewidth}{!}{
  \begin{tabular}{llcccc}
    \toprule
    \textbf{Distractions} & \textbf{Method} & \textbf{Avg. SR / Loc-SR} & \textbf{Open Drawer} & \textbf{Close Laptop} & \textbf{Close Microwave} \\
    \midrule
      & DP3          & 33.3 / 55.5 & 20.0 / 40.0 & 40.0 / 53.3 & 40.0 / 73.3 \\
      & Seer         & 37.8 / 64.4 & 26.7 / 53.3 & 46.7 / 73.3 & 40.0 / 66.7 \\
      \rowcolor{gray!15}
      \cellcolor{white!100}
      &  Ours         & 66.7 / 80.0 & 46.7 / 60.0 & 73.3 / \textbf{86.7} & \textbf{80.0} / \textbf{93.3} \\
      \rowcolor{gray!15}
     \cellcolor{white!100}
     \multirow{-4}{*}{None}
      &  Ours (w/ video) & \textbf{73.3} / \textbf{82.2} & \textbf{60.0} / \textbf{66.7} & \textbf{80.0} / \textbf{86.7} & \textbf{80.0} / \textbf{93.3} \\
    \midrule
    
      & DP3          & 28.9 / 57.8 & 20.0 / 46.7 & 40.0 / 46.7 & 26.7 / 80.0 \\
      & Seer         & 26.7 / 51.1 & 20.0 / 40.0 & 33.3 / 53.3 & 26.7 / 60.0 \\
      \rowcolor{gray!15}
      \cellcolor{white!100}
      &  Ours         & 62.2 / 77.8 & 40.0 / 53.3 & 66.7 / \textbf{86.7} & 80.0 / 93.3 \\
      \rowcolor{gray!15}
      \cellcolor{white!100}
      \multirow{-4}{*}{Background}
      &  Ours (w/ video) & \textbf{73.3} / \textbf{84.5} & \textbf{53.3} / \textbf{66.7} & \textbf{80.0} / \textbf{86.7} & \textbf{86.7} / \textbf{100.0} \\
    \midrule
    
      & DP3          & 24.4 / 42.2 & 20.0 / 33.3 & 20.0 / 53.3 & 33.3 / 40.0 \\
      & Seer         & 31.1 / 53.3 & 20.0 / 40.0 & 40.0 / 66.7 & 33.3 / 53.3 \\
      \rowcolor{gray!15}
      \cellcolor{white!100}
      &  Ours         & 62.2 / 71.1 & 40.0 / \textbf{60.0} & 80.0 / 80.0 & 66.7 / 73.3 \\
      \rowcolor{gray!15}
      \cellcolor{white!100}
      \multirow{-4}{*}{Object}
      &  Ours (w/ video) & \textbf{75.6} / \textbf{77.8} & \textbf{60.0} / \textbf{60.0} & \textbf{86.7} / \textbf{86.7} & \textbf{80.0} / \textbf{86.7} \\
    \bottomrule
  \end{tabular}
  }
  \vspace{-3mm}
\end{table}

Table~\ref{tab:real_exper} reports results across three generalization settings. Our method without video co-training achieves an average SR of 66.7\% (None), 62.2\% (Background), and 62.2\% (Object), substantially outperforming baselines under all conditions. The advantage holds for both SR and Loc-SR, indicating that KAI improves not only task completion but also contact-point localization. Notably, our method maintains stable performance across settings. From None to Object distractions, the average SR of our method without video drops by only 4.5 percentage points. In contrast, DP3 declines from 33.3\% to 24.4\% under object distractors, and Seer drops from 37.8\% to 26.7\% under background distractions, where its 2D visual prediction objective makes it more sensitive to background changes. This stability suggests that the kinematic priors embedded in KAI reduce the policy's reliance on surface-level visual correlations, yielding more robust behavior under distribution shifts.

Co-training with human interaction videos yields consistent improvements over our base method: +6.6\% (None), +11.1\% (Background), and +13.4\% (Object). These gains are substantially larger than those observed in simulation (+1.7\%), consistent with our earlier analysis that the perceptual domain gap between human videos and synthetic observations limits transfer in simulation. In the real world, human videos and robot observations share similar camera characteristics and scene statistics, allowing the kinematic knowledge in human demonstrations to transfer more effectively. The gains also grow with the severity of visual distraction, suggesting that exposure to diverse human interaction scenes helps the model localize contact points under challenging visual conditions. We also evaluate generalization to unseen object instances, with results reported in Appendix~\ref{sec:unseen_instances}.

\vspace{-3mm}

\section{Conclusion}

\vspace{-4mm}

We introduced KAI, a structured intermediate representation that embeds kinematic priors into policy learning for articulated object manipulation. By encoding the state and motion of articulated parts, KAI provides a geometric inductive bias improving both sample efficiency and generalization. In simulation, our method achieves an average success rate of 82.9\%, matching or surpassing baseline performance with half of the demonstration data, and these gains are particularly pronounced in low-data regimes. In the real world, policies trained in a single clean setting transfer robustly to unseen backgrounds and distractors. KAI's action-agnostic design further enables co-training with human interaction videos, yielding over 70\% average success rate under visual distractions.

\vspace{-3mm}

\section{Limitations}
\label{sec:limitations} 

\vspace{-4mm}

Two aspects of the current work point to natural directions for future investigation. First, our evaluation focuses on short-horizon manipulation tasks. Extending KAI to long-horizon sequential tasks is a natural direction for future work. Second, the real-world experiments use a single robot platform. Validating the approach across different embodiments would further characterize its generality.



\clearpage


\bibliography{references}  

@inproceedings{jain2021screwnet,
  title={Screwnet: Category-independent articulation model estimation from depth images using screw theory},
  author={Jain, Ajinkya and Lioutikov, Rudolf and Chuck, Caleb and Niekum, Scott},
  booktitle={2021 IEEE International Conference on Robotics and Automation (ICRA)},
  pages={13670--13677},
  year={2021},
  organization={IEEE}
}

@inproceedings{jain2022distributional,
  title={Distributional depth-based estimation of object articulation models},
  author={Jain, Ajinkya and Giguere, Stephen and Lioutikov, Rudolf and Niekum, Scott},
  booktitle={Conference on Robot Learning},
  pages={1611--1621},
  year={2022},
  organization={PMLR}
}

@inproceedings{zeng2021visual,
  title={Visual identification of articulated object parts},
  author={Zeng, Vicky and Lee, Tabitha Edith and Liang, Jacky and Kroemer, Oliver},
  booktitle={2021 IEEE/RSJ International Conference on Intelligent Robots and Systems (IROS)},
  pages={2443--2450},
  year={2021},
  organization={IEEE}
}

@inproceedings{li2020category,
  title={Category-level articulated object pose estimation},
  author={Li, Xiaolong and Wang, He and Yi, Li and Guibas, Leonidas J and Abbott, A Lynn and Song, Shuran},
  booktitle={Proceedings of the IEEE/CVF conference on computer vision and pattern recognition},
  pages={3706--3715},
  year={2020}
}

@inproceedings{yu2024gamma,
  title={Gamma: Generalizable articulation modeling and manipulation for articulated objects},
  author={Yu, Qiaojun and Wang, Junbo and Liu, Wenhai and Hao, Ce and Liu, Liu and Shao, Lin and Wang, Weiming and Lu, Cewu},
  booktitle={2024 IEEE International Conference on Robotics and Automation (ICRA)},
  pages={5419--5426},
  year={2024},
  organization={IEEE}
}

@inproceedings{geng2023gapartnet,
  title={Gapartnet: Cross-category domain-generalizable object perception and manipulation via generalizable and actionable parts},
  author={Geng, Haoran and Xu, Helin and Zhao, Chengyang and Xu, Chao and Yi, Li and Huang, Siyuan and Wang, He},
  booktitle={Proceedings of the IEEE/CVF Conference on Computer Vision and Pattern Recognition},
  pages={7081--7091},
  year={2023}
}

@inproceedings{wang2024rpmart,
  title={Rpmart: Towards robust perception and manipulation for articulated objects},
  author={Wang, Junbo and Liu, Wenhai and Yu, Qiaojun and You, Yang and Liu, Liu and Wang, Weiming and Lu, Cewu},
  booktitle={2024 IEEE/RSJ International Conference on Intelligent Robots and Systems (IROS)},
  pages={7270--7277},
  year={2024},
  organization={IEEE}
}

@inproceedings{morlans2024grasp,
  title={Ao-grasp: Articulated object grasp generation},
  author={Morlans, Carlota Par{\'e}s and Chen, Claire and Weng, Yijia and Yi, Michelle and Huang, Yuying and Heppert, Nick and Zhou, Linqi and Guibas, Leonidas and Bohg, Jeannette},
  booktitle={2024 IEEE/RSJ International Conference on Intelligent Robots and Systems (IROS)},
  pages={13096--13103},
  year={2024},
  organization={IEEE}
}

@article{eisner2022flowbot3d,
  title={Flowbot3d: Learning 3d articulation flow to manipulate articulated objects},
  author={Eisner, Ben and Zhang, Harry and Held, David},
  journal={arXiv preprint arXiv:2205.04382},
  year={2022}
}

@article{zhang2023flowbot++,
  title={Flowbot++: Learning generalized articulated objects manipulation via articulation projection},
  author={Zhang, Harry and Eisner, Ben and Held, David},
  journal={arXiv preprint arXiv:2306.12893},
  year={2023}
}

@inproceedings{cui2025gapartmanip,
  title={Gapartmanip: A large-scale part-centric dataset for material-agnostic articulated object manipulation},
  author={Cui, Wenbo and Zhao, Chengyang and Wei, Songlin and Zhang, Jiazhao and Geng, Haoran and Chen, Yaran and Li, Haoran and Wang, He},
  booktitle={2025 IEEE International Conference on Robotics and Automation (ICRA)},
  pages={14791--14798},
  year={2025},
  organization={IEEE}
}

@article{wang2024articulated,
  title={Articulated object manipulation using online axis estimation with sam2-based tracking},
  author={Wang, Xi and Chen, Tianxing and Yu, Qiaojun and Xu, Tianling and Chen, Zanxin and Fu, Yiting and He, Ziqi and Lu, Cewu and Mu, Yao and Luo, Ping},
  journal={arXiv preprint arXiv:2409.16287},
  year={2024}
}

@inproceedings{mo2021where2act,
  title={Where2act: From pixels to actions for articulated 3d objects},
  author={Mo, Kaichun and Guibas, Leonidas J and Mukadam, Mustafa and Gupta, Abhinav and Tulsiani, Shubham},
  booktitle={Proceedings of the IEEE/CVF International Conference on Computer Vision},
  pages={6813--6823},
  year={2021}
}

@article{wu2021vat,
  title={Vat-mart: Learning visual action trajectory proposals for manipulating 3d articulated objects},
  author={Wu, Ruihai and Zhao, Yan and Mo, Kaichun and Guo, Zizheng and Wang, Yian and Wu, Tianhao and Fan, Qingnan and Chen, Xuelin and Guibas, Leonidas and Dong, Hao},
  journal={arXiv preprint arXiv:2106.14440},
  year={2021}
}

@article{xu2022universal,
  title={Universal manipulation policy network for articulated objects},
  author={Xu, Zhenjia and He, Zhanpeng and Song, Shuran},
  journal={IEEE robotics and automation letters},
  volume={7},
  number={2},
  pages={2447--2454},
  year={2022},
  publisher={IEEE}
}

@article{wang2025articubot,
  title={ArticuBot: Learning Universal Articulated Object Manipulation Policy via Large Scale Simulation},
  author={Wang, Yufei and Wang, Ziyu and Nakura, Mino and Bhowal, Pratik and Kuo, Chia-Liang and Chen, Yi-Ting and Erickson, Zackory and Held, David},
  journal={arXiv preprint arXiv:2503.03045},
  year={2025}
}

@inproceedings{geng2023rlafford,
  title={Rlafford: End-to-end affordance learning for robotic manipulation},
  author={Geng, Yiran and An, Boshi and Geng, Haoran and Chen, Yuanpei and Yang, Yaodong and Dong, Hao},
  booktitle={2023 IEEE International conference on robotics and automation (ICRA)},
  pages={5880--5886},
  year={2023},
  organization={IEEE}
}

@article{ning2023where2explore,
  title={Where2explore: Few-shot affordance learning for unseen novel categories of articulated objects},
  author={Ning, Chuanruo and Wu, Ruihai and Lu, Haoran and Mo, Kaichun and Dong, Hao},
  journal={Advances in Neural Information Processing Systems},
  volume={36},
  pages={4585--4596},
  year={2023}
}

@inproceedings{wang2022adaafford,
  title={Adaafford: Learning to adapt manipulation affordance for 3d articulated objects via few-shot interactions},
  author={Wang, Yian and Wu, Ruihai and Mo, Kaichun and Ke, Jiaqi and Fan, Qingnan and Guibas, Leonidas J and Dong, Hao},
  booktitle={European conference on computer vision},
  pages={90--107},
  year={2022},
  organization={Springer}
}

@inproceedings{yu2025uniaff,
  title={Uniaff: A unified representation of affordances for tool usage and articulation with vision-language models},
  author={Yu, Qiaojun and Huang, Siyuan and Yuan, Xibin and Jiang, Zhengkai and Hao, Ce and Li, Xin and Chang, Haonan and Wang, Junbo and Liu, Liu and Li, Hongsheng and others},
  booktitle={2025 IEEE International Conference on Robotics and Automation (ICRA)},
  pages={8980--8987},
  year={2025},
  organization={IEEE}
}

@article{fang2025airexo,
  title={AirExo-2: Scaling up Generalizable Robotic Imitation Learning with Low-Cost Exoskeletons},
  author={Fang, Hongjie and Wang, Chenxi and Wang, Yiming and Chen, Jingjing and Xia, Shangning and Lv, Jun and He, Zihao and Yi, Xiyan and Guo, Yunhan and Zhan, Xinyu and others},
  journal={arXiv preprint arXiv:2503.03081},
  year={2025}
}

@article{oquab2023dinov2,
  title={Dinov2: Learning robust visual features without supervision},
  author={Oquab, Maxime and Darcet, Timoth{\'e}e and Moutakanni, Th{\'e}o and Vo, Huy and Szafraniec, Marc and Khalidov, Vasil and Fernandez, Pierre and Haziza, Daniel and Massa, Francisco and El-Nouby, Alaaeldin and others},
  journal={arXiv preprint arXiv:2304.07193},
  year={2023}
}

@inproceedings{radford2021learning,
  title={Learning transferable visual models from natural language supervision},
  author={Radford, Alec and Kim, Jong Wook and Hallacy, Chris and Ramesh, Aditya and Goh, Gabriel and Agarwal, Sandhini and Sastry, Girish and Askell, Amanda and Mishkin, Pamela and Clark, Jack and others},
  booktitle={International conference on machine learning},
  pages={8748--8763},
  year={2021},
  organization={PmLR}
}

@inproceedings{liu2022hoi4d,
  title={Hoi4d: A 4d egocentric dataset for category-level human-object interaction},
  author={Liu, Yunze and Liu, Yun and Jiang, Che and Lyu, Kangbo and Wan, Weikang and Shen, Hao and Liang, Boqiang and Fu, Zhoujie and Wang, He and Yi, Li},
  booktitle={Proceedings of the IEEE/CVF Conference on Computer Vision and Pattern Recognition},
  pages={21013--21022},
  year={2022}
}

@software{NVIDIA_Isaac_Sim,
author = {{NVIDIA}},
license = {Apache-2.0},
title = {{Isaac Sim}},
url = {https://github.com/isaac-sim/IsaacSim},
version = {5.1.0}
}

@article{zhao2023learning,
  title={Learning fine-grained bimanual manipulation with low-cost hardware},
  author={Zhao, Tony Z and Kumar, Vikash and Levine, Sergey and Finn, Chelsea},
  journal={arXiv preprint arXiv:2304.13705},
  year={2023}
}

@article{tian2024predictive,
  title={Predictive inverse dynamics models are scalable learners for robotic manipulation},
  author={Tian, Yang and Yang, Sizhe and Zeng, Jia and Wang, Ping and Lin, Dahua and Dong, Hao and Pang, Jiangmiao},
  journal={arXiv preprint arXiv:2412.15109},
  year={2024}
}

@article{ze20243d,
  title={3d diffusion policy: Generalizable visuomotor policy learning via simple 3d representations},
  author={Ze, Yanjie and Zhang, Gu and Zhang, Kangning and Hu, Chenyuan and Wang, Muhan and Xu, Huazhe},
  journal={arXiv preprint arXiv:2403.03954},
  year={2024}
}

\clearpage
\appendix
\section{Appendix}
\subsection{Implementation Details}
\label{sec:implementation_details}
\textbf{Vision.} At each timestep, visual inputs are captured from two viewpoints: an eye-on-hand camera and an eye-on-base camera. For semantic feature extraction, we employ a pre-trained DINOv2-base model to independently process both RGB images, generating feature maps with 128 output channels. These 2D semantic features are then lifted into 3D space through coordinate transformation and subsequently concatenated.
For geometric feature extraction, we back-project the depth images from both cameras into 3D space to form point clouds. Each point cloud undergoes Farthest Point Sampling to select 512 representative points, resulting in a combined point cloud of 1024 points containing both coordinate and color information. This unified point cloud is then processed by a ResNet14-based network to extract geometric features, also with an output channel dimension of 128.
All 3D coordinates are defined in the world coordinate system, with the robot base position manually calibrated for real-world experiments. The spatial aligner module subsequently fuses the 2D semantic features and 3D geometric features, producing an output of 14 tokens with 512 channels each. Finally, an MLP projects this representation to 384 dimensions, serving as the final visual embedding.

\textbf{Robot State.}
The robot state representation comprises two distinct components: arm state and gripper state. For the arm state, we employ joint positions (qpos) in simulation experiments, while in real-world experiments we use a 6-dimensional representation consisting of end-effector position and rotation in Euler angles. The gripper state is represented as a binary value indicating open or closed status.
We tokenize the robot state using a dedicated MLP encoder. The processing pipeline begins by converting the binary gripper state into a one-hot encoding. Both the one-hot gripper encoding and the continuous arm state are then processed through separate linear layers. The resulting feature vectors are concatenated and passed through a final linear projection layer to generate the state token, which integrates both proprioceptive components into a unified representation for downstream processing.

\textbf{Language Instruction.}
Natural language instructions are processed using the CLIP ViT-B/32 text encoder \cite{radford2021learning} to extract semantic representations. The resulting text embeddings are subsequently projected through a linear layer to generate the final language token.

\textbf{Transformer Backbone.}
Our model employs a 24-layer GPT-2 transformer architecture with a hidden size of 384 as the core processing backbone. The input to the transformer consists of a comprehensive sequence combining the visual embeddings, robot state tokens, language instruction features, along with dedicated readout tokens for predicting both the Kinematic-Aware Articulation Interface (KAI) and robot actions.

\textbf{Decoder.}
Our decoding framework consists of two specialized branches for KAI and action prediction. For the KAI output, the KAI readout token from the transformer blocks is processed by the KAI decoder, which first extracts shared features through an MLP, then branches into two dedicated MLP decoders: a location decoder for keypoint positions and a motion decoder for displacement trajectories. Similarly, for action prediction, the action readout token is fed into the action decoder that first employs an MLP to extract shared representations, then diverges into two separate MLP decoders: an arm action decoder for continuous motion control and a gripper action decoder for discrete grasping commands.

\textbf{Training.} Our model comprises 304M parameters, of which 77M are tunable. The policy was trained for approximately 60 epochs, including a 5-epoch warmup phase, over about 8 hours using 32 NVIDIA RTX 4090 GPUs. Relevant hyperparameters are detailed in Table~\ref{tab:hyperparameters}.

\begin{table}[h]
\centering
\caption{\textbf{Hyperparameter Settings.}}
\resizebox{0.5\linewidth}{!}{
\begin{tabular}{cc}
\toprule
Hyperparameter & Value \\
\midrule
Learning rate schedule & Cosine decay \\
Learning rate & 0.001 \\
Optimizer & AdamW \\
Batch size & 512 \\
Hidden dimension & 384 \\
Location loss weight $\alpha_1$ & 1.0 \\
Motion loss weight $\alpha_2$ & 1.0 \\
Geometric loss weight $\alpha_3$ & 0.001 \\
KAI loss weight $\lambda_1$ & 1000.0 \\
Action loss weight $\lambda_2$ & 1.0 \\
\bottomrule
\end{tabular}
}
\label{tab:hyperparameters}
\end{table}

\textbf{Inference.} Simulation experiments are evaluated on NVIDIA RTX 4090 GPUs, while real-world deployment runs on an NVIDIA RTX 4080 SUPER GPU.

\subsection{Baseline Implementation}
\label{sec:baseline_implementation}
In our simulation experiments, we evaluate and report results for five baseline methods: ACT \cite{zhao2023learning}, Seer \cite{tian2024predictive}, DP3 \cite{ze20243d}, RISE-2 \cite{fang2025airexo}, and ArticuBot \cite{wang2025articubot}. 
For ACT, we report the performance of the best checkpoint that achieved the minimum validation loss during training. 
For Seer, we average the success rates across three checkpoints (epochs 30, 34, and 39) after training convergence to ensure stable performance measurement. 
For DP3, we employ a horizon of 16 and observation steps of 8, with point cloud processing pipelines consistent with our method to ensure fair comparison. 
For RISE-2, we adapt the original implementation by modifying the output action dimension to suit single-arm manipulation, and report results obtained with a batch size of 256 and learning rate of 0.001 after full training convergence. 
For ArticuBot, we adapt the original method to accommodate close-type tasks—which lack explicit grasping phases—by redefining the sub-goal as the end-effector position at the final timestep of a sliding window (size=16), rather than using the grasping-time position specified in the original implementation.
All baselines were trained and evaluated under identical environmental conditions and task definitions.

\subsection{Data Efficiency Details}
\label{sec:data_efficiency_details}

\begin{table*}[htbp]
\centering
\caption{\textbf{Detailed Data Efficiency Analysis.} Extended results showing task-wise performance across different data volumes.}
\label{tab:data_efficiency}
\small
\resizebox{\linewidth}{!}{
\begin{tabular}{@{}cc|ccccccc@{}}
\toprule
 Method & Data Volume & Avg. SR & Open Drawer & Open Door & Open Laptop & Close Drawer & Close Door & Close Laptop \\

\midrule

\multirow{4}{*}{DP3}
 & 100 & 20.1 & 12.0 & 5.5 & 1.0 & 11.0 & 1.5 & 89.5 \\
 & 200 & 13.0 & 1.0 & 4.0 & 4.5 & 9.5 & 4.5 & 54.5 \\
 & 400 & 24.3 & 24.0 & 8.0 & 10.5 & 0.0 & 8.5 & 94.5 \\
 & 800 & 37.3 & 25.0 & 12.5 & 25.5 & 2.0 & 97.0 & 62.0 \\
\midrule
\multirow{4}{*}{Seer}
 & 100 & 20.2 & 7.0 & 17.5 & 7.7 & 12.5 & 27.0 & 49.7 \\
 & 200 & 46.5 & 52.7 & 41.8 & 36.8 & 25.0 & 49.2 & 73.7 \\
 & 400 & 60.4 & 63.2 & 61.8 & 45.3 & 49.2 & 64.0 & 78.7 \\
 & 800 & 73.4 & 84.8 & 67.2 & 67.7 & 68.1 & 63.7 & 89.2 \\
\midrule
\multirow{4}{*}{Ours}
 & 100 & 42.4 & 31.4 & 21.2 & 48.7 & 28.5 & 40.2 & 84.7 \\
 & 200 & 62.6 & 54.8 & 35.8 & 74.5 & 40.8 & 72.2 & 97.5 \\
 & 400 & 75.9 & 73.8 & 54.7 & 84.7 & 59.5 & 84.5 & 98.0 \\
 & 800 & 82.9 & 86.7 & 54.5 & 81.7 & 82.3 & 92.5 & 100.0 \\
\midrule
\multirow{4}{*}{Ours (w/ video)}
 & 100 & 47.6 & 29.0 & 13.2 & 67.7 & 32.0 & 40.3 & 99.2 \\
 & 200 & 61.7 & 53.1 & 32.5 & 79.3 & 44.5 & 70.5 & 100.0 \\
 & 400 & 76.1 & 82.7 & 43.2 & 91.0 & 60.8 & 85.7 & 100.0 \\
 & 800 & 84.6 & 86.8 & 66.0 & 85.0 & 81.8 & 88.8 & 99.2 \\

\bottomrule
\end{tabular}
}
\end{table*} 

Table~\ref{tab:data_efficiency} reports task-wise success rates across four data volumes for our method and two baselines. Our method consistently outperforms both DP3 and Seer at every data scale, with the advantage most pronounced in the low-data regime. Notably, with only 100 demonstrations, our method without and with videos co-training attains an average success rate of 42.4\% and 47.6\%, respectively, more than doubling that of DP3 (20.1\%) and Seer (20.2\%).

\subsection{Quantitative Analysis of KAI Prediction}
\label{sec:kai_mde_analysis}

\begin{table*}[htbp]
  \centering
  \caption{\textbf{Quantitative evaluation of task success and KAI location accuracy.} We report the Success Rate (SR, \%) and the Mean Distance Error (MDE, cm), which measures the mean Euclidean distance between the predicted KAI locations and the ground truth (GT).}
  \label{tab:kai_mde_results}
  \resizebox{\linewidth}{!}{
  \begin{tabular}{cccccccc}
    \toprule
    \multirow{2}{*}{Method} & \multicolumn{7}{c}{SR (\%) $\uparrow$ / MDE (cm) $\downarrow$} \\
    \cmidrule(lr){2-8} 
    & Average & Open Drawer & Open Door & Open Laptop & Close Drawer & Close Door & Close Laptop \\
    \midrule \midrule
    Ours & 82.9 / 2.65 & 86.7 / 2.81 & 54.5 / 4.61 & 81.7 / \textbf{1.95} & \textbf{82.3} / \textbf{4.02} & \textbf{92.5} / \textbf{1.18} & \textbf{100.0} / \textbf{1.30} \\
    Ours (w/ video) & \textbf{84.6} / \textbf{2.59} & \textbf{86.8} / \textbf{2.69} & \textbf{66.0} / \textbf{4.00} & \textbf{85.0} / 2.10 & 81.8 / \textbf{4.02} & 88.8 / 1.38 & 99.2 / 1.40 \\
    \bottomrule
  \end{tabular}
  }
\end{table*}

 To further investigate the precision of the KAI, we report the Mean Distance Error (MDE) in Table~\ref{tab:kai_mde_results}. The MDE quantifies the $L_2$ distance between the predicted locations $\{ \hat{\mathbf{l}}^i \}_{i=1}^K$ and the ground-truth locations $\{ \mathbf{l}^i \}_{i=1}^K$. Specifically, it is calculated as the average Euclidean distance across all $K$ keypoints over all $T$ timesteps and $M$ evaluation trials:
\begin{equation}
    \text{MDE} = \frac{1}{M \cdot T \cdot K} \sum_{m=1}^{M} \sum_{t=1}^{T} \sum_{i=1}^{K} \left\| \hat{\mathbf{l}}^i_{m,t} - \mathbf{l}^i_{m,t} \right\|_2.
\end{equation}
 As demonstrated in the results, we observe a clear negative correlation between Success Rate (SR) and MDE: generally, a lower MDE indicates more precise localization, which directly facilitates higher task success. Notably, while co-training with human videos improves the average performance, certain tasks exhibit a slight increase in MDE and a corresponding drop in success rate after co-training. This phenomenon likely stems from the domain gap between real-world video distributions and simulated task environments, where visual gap and 3D sensor noise from the wild can hinder the precise coordinate prediction within the simulator.

\subsection{Evaluation under Seen Backgrounds and Lighting}
\label{sec:seen_bg_light}

\begin{table}[htbp]
  \centering
  \vspace{-10pt}
  \caption{\textbf{Evaluation under seen backgrounds and lighting.}}
  \label{tab:seen_bg_light}
  \vspace{2mm}
  \begin{tabular}{lcccc}
    \toprule
    \textbf{Method} & \textbf{Avg. SR (\%)} & \textbf{Open Drawer} & \textbf{Close Door} & \textbf{Close Laptop} \\
    \midrule
    ACT & 52.5 & 42.5 & 57.5 & 57.5 \\
    Seer & 78.8 & 88.5 & 66.2 & 81.8 \\
    \rowcolor{gray!15} Ours & 90.6 & 86.5 & 90.3 & 95.0 \\
    \bottomrule
  \end{tabular}
  \vspace{-5pt}
\end{table}

Table~\ref{tab:seen_bg_light} reports simulation results under the same backgrounds and lighting used during training, evaluated on three representative tasks. KAI maintains a clear margin over both baselines, confirming that its performance advantage stems from kinematic reasoning rather than from implicit robustness to visual distribution shifts inherited from its perception backbone.

\subsection{Visualizations of Generalization Settings in Real-world Tasks}
\label{sec:visualizations_generalization}
\begin{figure*}[h]
    \centering
    \begin{subfigure}[b]{\textwidth}
        \centering
        \includegraphics[width=0.19\textwidth]{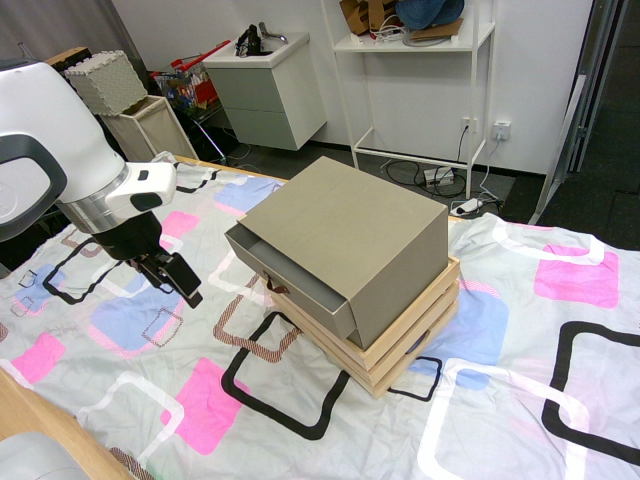}\hfill
        \includegraphics[width=0.19\textwidth]{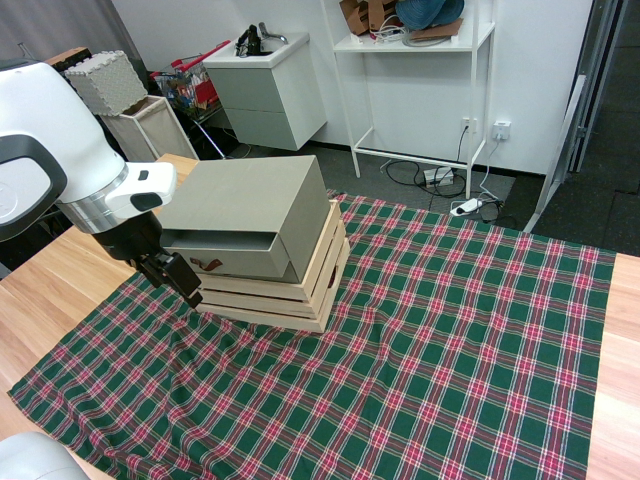}\hfill
        \includegraphics[width=0.19\textwidth]{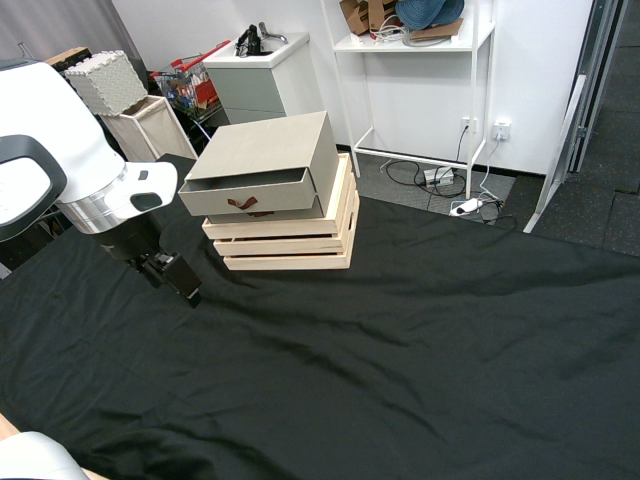}\hfill
        \includegraphics[width=0.19\textwidth]{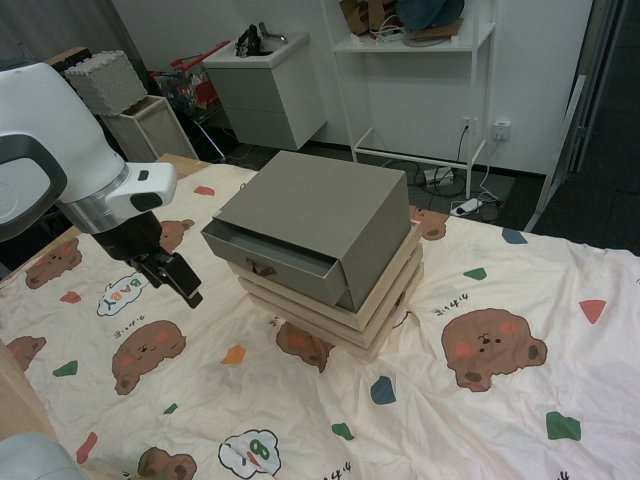}\hfill
        \includegraphics[width=0.19\textwidth]{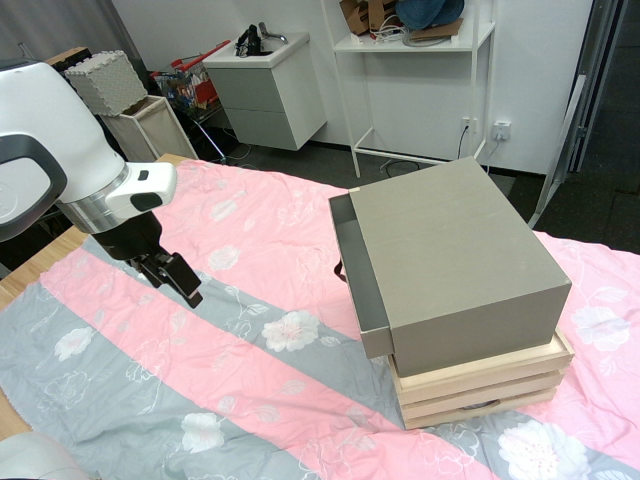}
        \caption{Open Drawer} 

    \end{subfigure}

    \vspace{1em} 

    \begin{subfigure}[b]{\textwidth}
        \centering
        \includegraphics[width=0.19\textwidth]{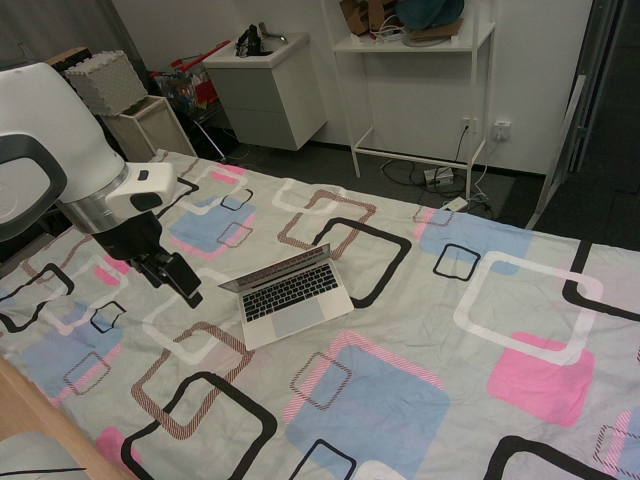}\hfill
        \includegraphics[width=0.19\textwidth]{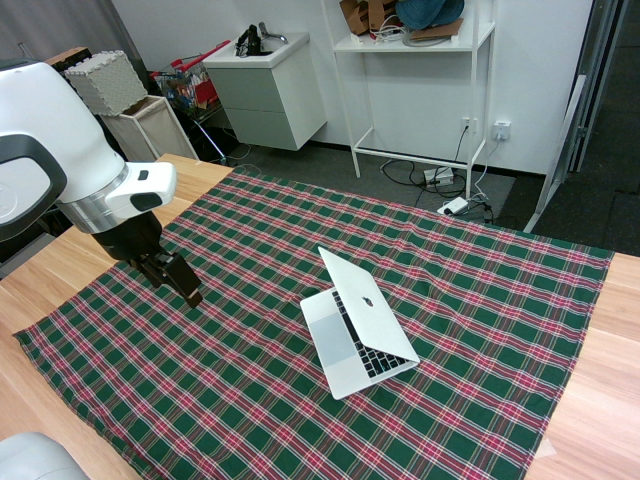}\hfill
        \includegraphics[width=0.19\textwidth]{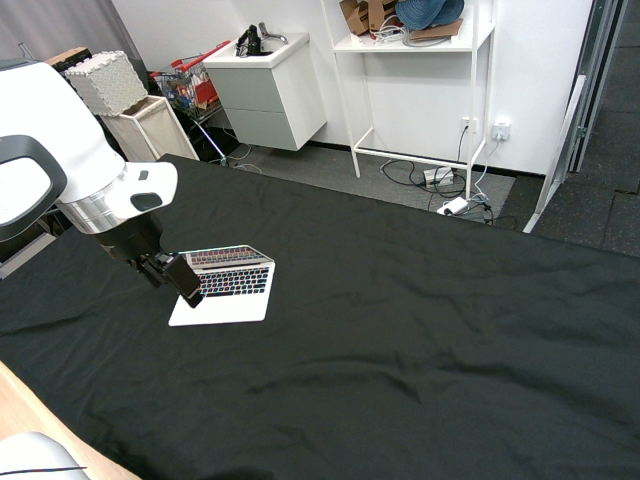}\hfill
        \includegraphics[width=0.19\textwidth]{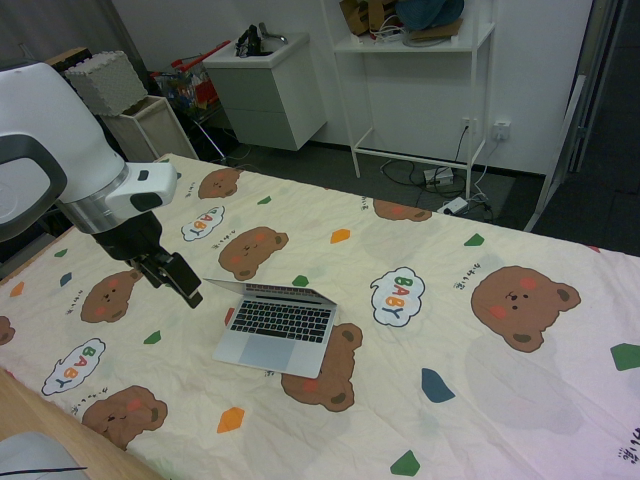}\hfill
        \includegraphics[width=0.19\textwidth]{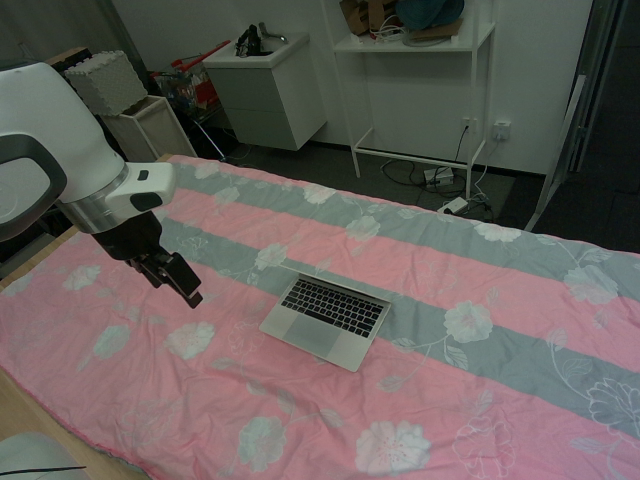}
        \caption{Close Laptop} 

    \end{subfigure}

    \vspace{1em} 

    \begin{subfigure}[b]{\textwidth}
        \centering
        \includegraphics[width=0.19\textwidth]{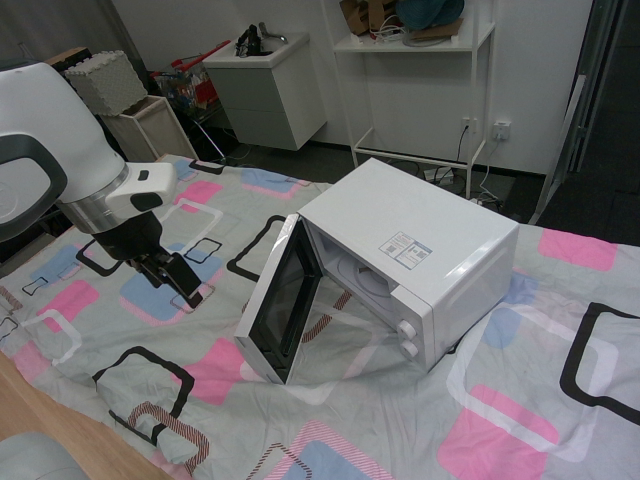}\hfill
        \includegraphics[width=0.19\textwidth]{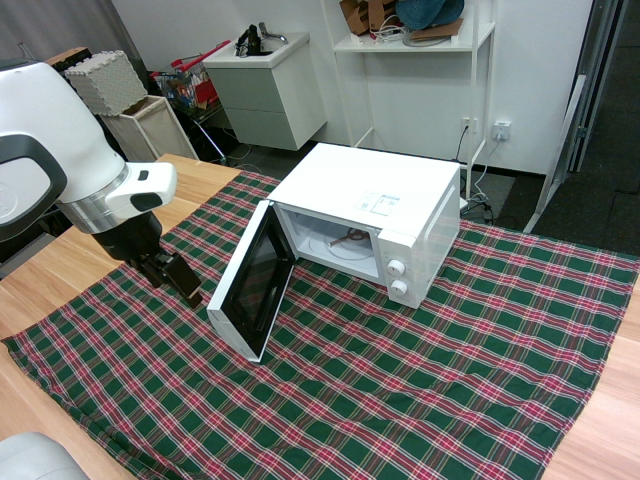}\hfill
        \includegraphics[width=0.19\textwidth]{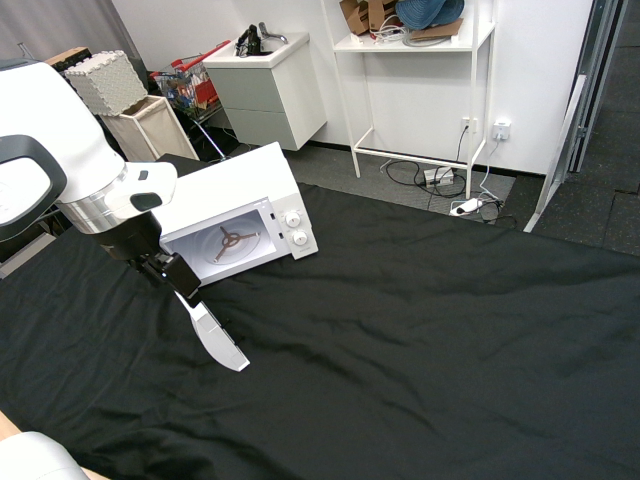}\hfill
        \includegraphics[width=0.19\textwidth]{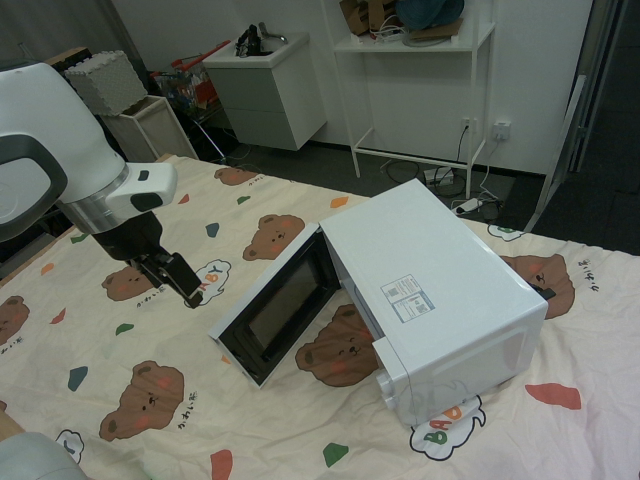}\hfill
        \includegraphics[width=0.19\textwidth]{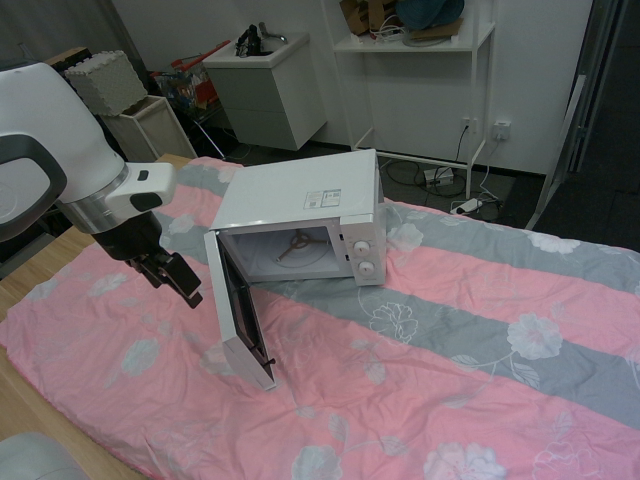}
        \caption{Close Microwave} 

    \end{subfigure}

    \caption{Visualization of five initial states per task under background distraction generalization setup.}
    \label{fig:background distractions Initial settings}
\end{figure*}

\begin{figure*}[h]
    \centering
    \begin{subfigure}[b]{\textwidth}
        \centering
        \includegraphics[width=0.19\textwidth]{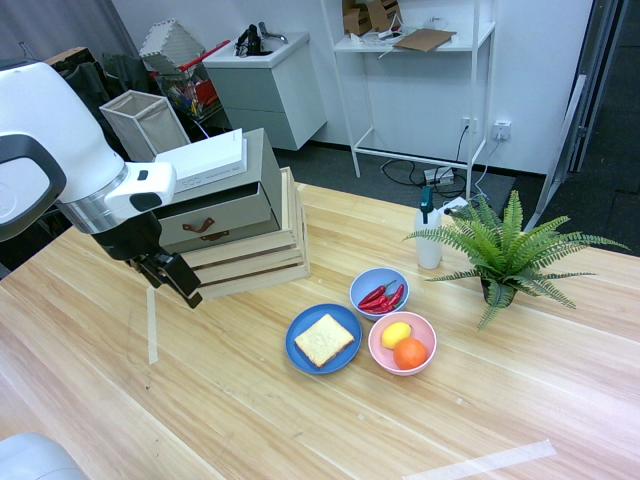}\hfill
        \includegraphics[width=0.19\textwidth]{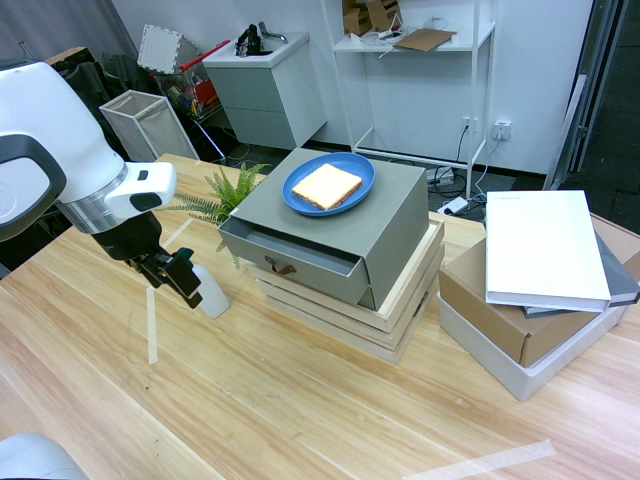}\hfill
        \includegraphics[width=0.19\textwidth]{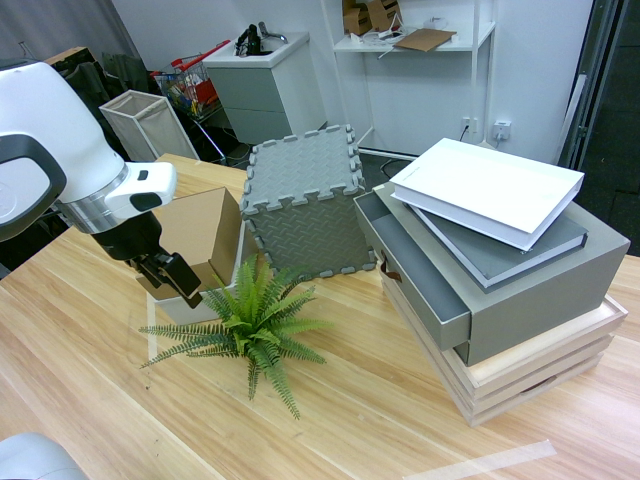}\hfill
        \includegraphics[width=0.19\textwidth]{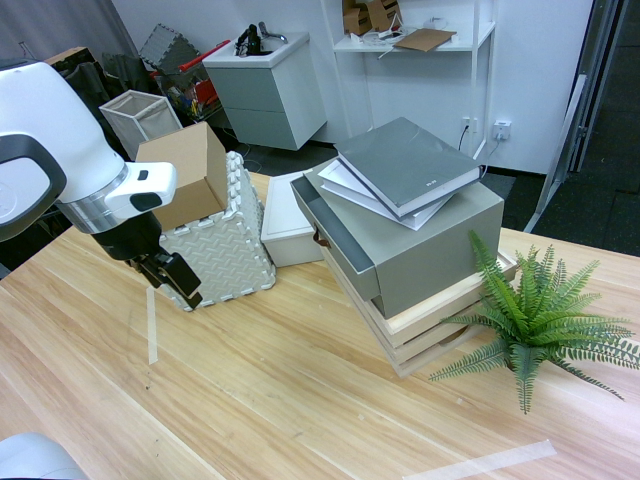}\hfill
        \includegraphics[width=0.19\textwidth]{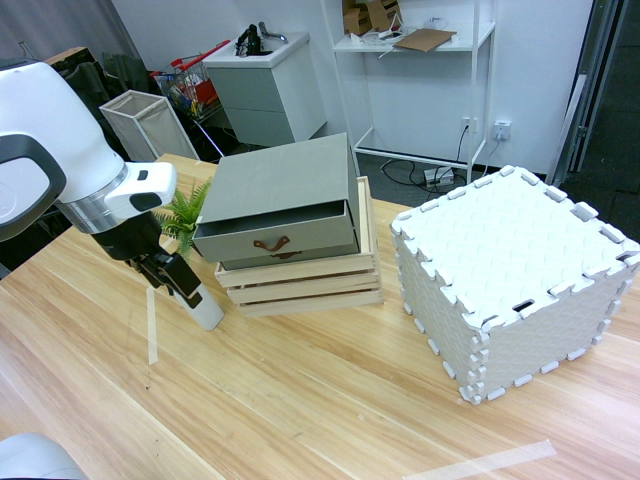}
        \caption{Open Drawer} 

    \end{subfigure}

    \vspace{1em} 

    \begin{subfigure}[b]{\textwidth}
        \centering
        \includegraphics[width=0.19\textwidth]{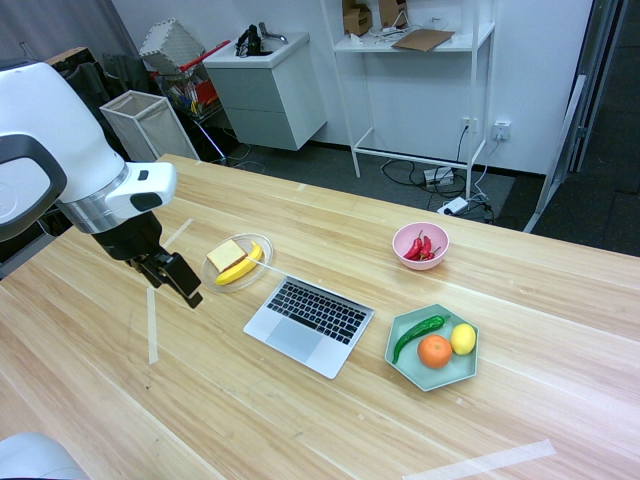}\hfill
        \includegraphics[width=0.19\textwidth]{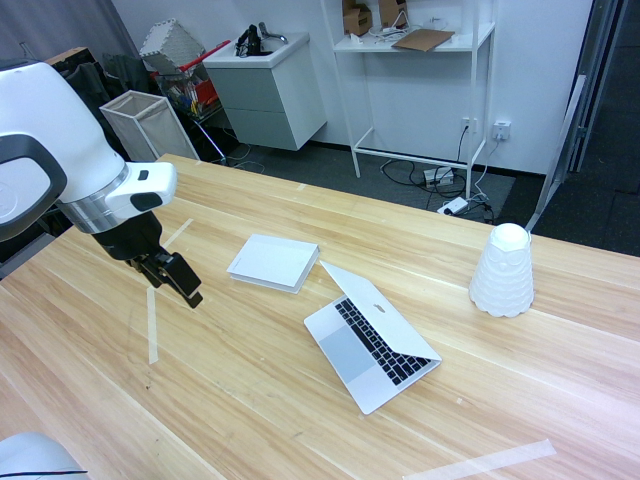}\hfill
        \includegraphics[width=0.19\textwidth]{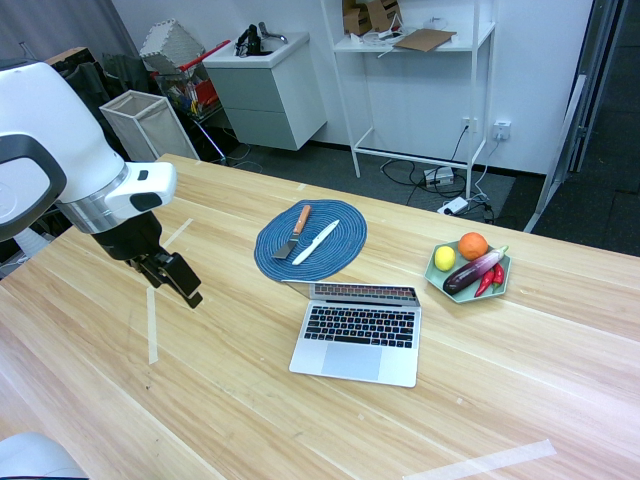}\hfill
        \includegraphics[width=0.19\textwidth]{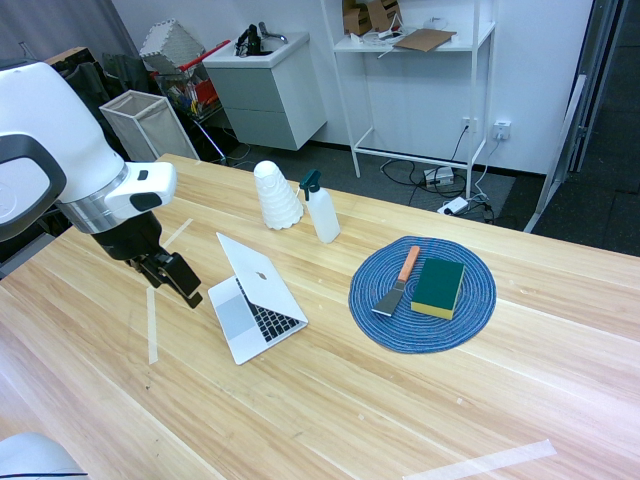}\hfill
        \includegraphics[width=0.19\textwidth]{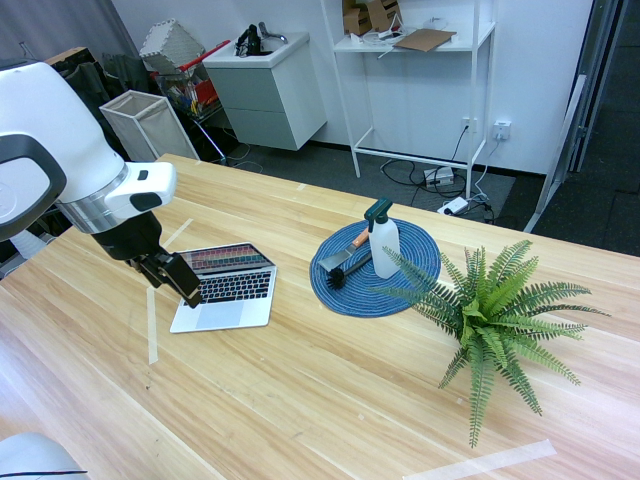}
        \caption{Close Laptop} 

    \end{subfigure}

    \vspace{1em} 

    \begin{subfigure}[b]{\textwidth}
        \centering
        \includegraphics[width=0.19\textwidth]{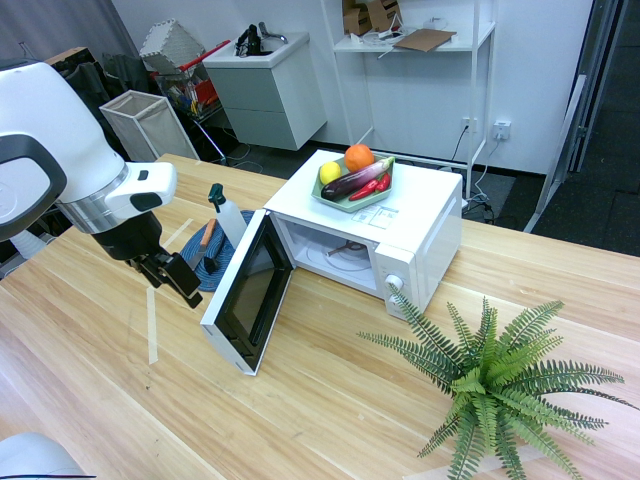}\hfill 
        \includegraphics[width=0.19\textwidth]{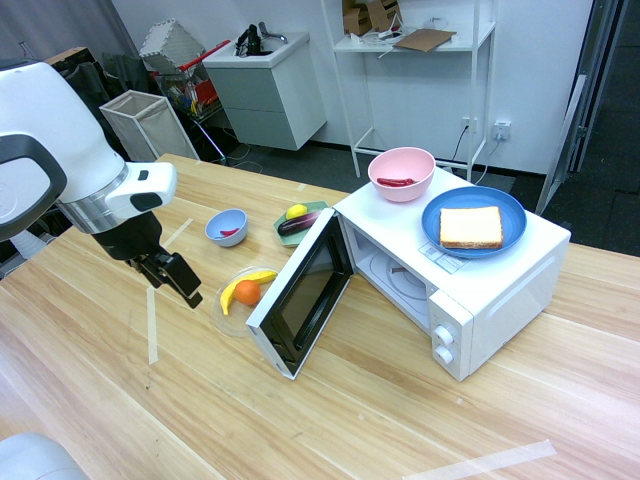}\hfill
        \includegraphics[width=0.19\textwidth]{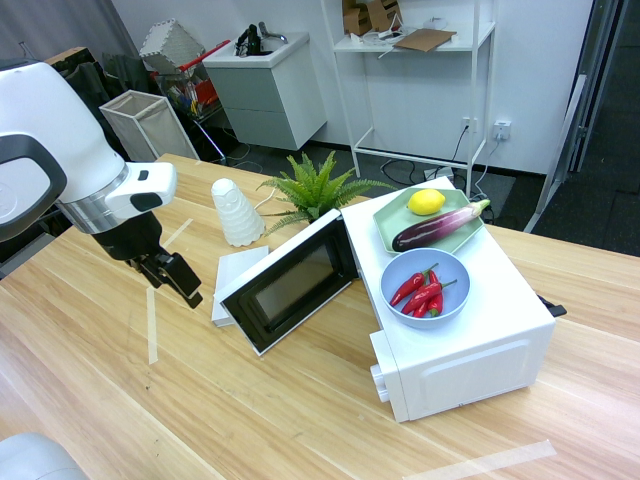}\hfill
        \includegraphics[width=0.19\textwidth]{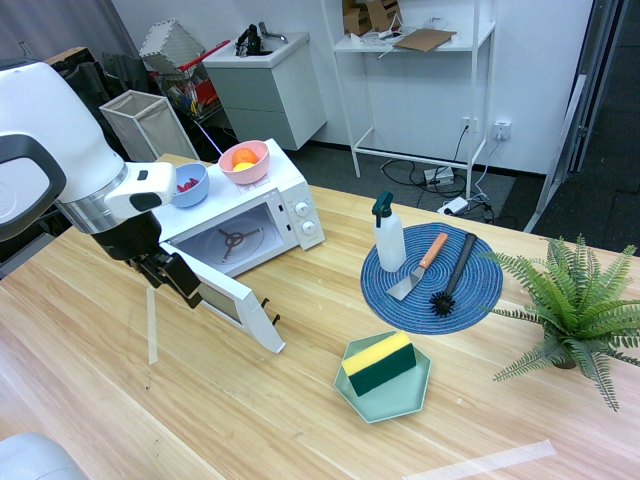}\hfill
        \includegraphics[width=0.19\textwidth]{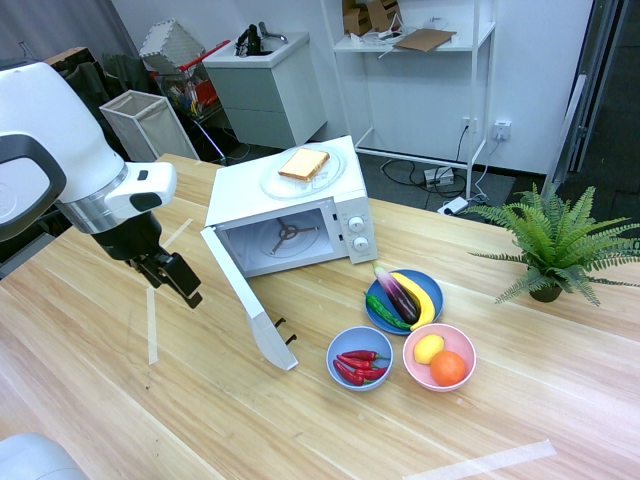}
        \caption{Close Microwave} 

    \end{subfigure}

    \caption{Visualization of five initial states per task under object distractor generalization setup.}
    \label{fig:object distractors Initial settings}
\end{figure*}

To assess policy robustness, we evaluate methods under two generalization setups: background distractions and object distractors. Figures~\ref{fig:background distractions Initial settings} and \ref{fig:object distractors Initial settings} illustrate five representative initial configurations for each task under varying table textures and diverse arrangements of the target and distractor objects, respectively. These visualizations demonstrate the range of environmental conditions and spatial layouts used to evaluate generalization.

\subsection{Generalization to Unseen Instances}
\label{sec:unseen_instances}

\begin{table}[htbp]
  \centering
  \begin{minipage}[t]{0.4\textwidth}
    \centering
    \caption{Unseen instance generalization in simulation.}
    \label{tab:sim_unseen}
    \vspace{2mm}
    \begin{tabular}{lc}
      \toprule
      \textbf{Method} & \textbf{SR (\%)} \\
      \midrule
      Seer & 18.2 \\
      \rowcolor{gray!15} Ours & 38.0 \\
      \bottomrule
    \end{tabular}
  \end{minipage}
  \hfill
  \begin{minipage}[t]{0.58\textwidth}
    \centering
    \caption{Real-world unseen instance generalization.}
    \label{tab:real_unseen}
    \vspace{2mm}
    \begin{tabular}{lc}
      \toprule
      \textbf{Method} & \textbf{SR / Loc-SR (\%)} \\
      \midrule
      DP3 & 46.7 / 53.3 \\
      Seer & 46.7 / 53.3 \\
      \midrule
      \rowcolor{gray!15} Ours & 66.7 / 73.3 \\
      \rowcolor{gray!15} Ours (w/ video) & 80.0 / 93.3 \\
      \bottomrule
    \end{tabular}
  \end{minipage}
\end{table}


To examine whether KAI learns transferable kinematic principles rather than memorizing object-specific geometries, we evaluate zero-shot transfer to unseen object instances. In simulation, we train on 32 laptop instances and test on 8 unseen instances, all in the close laptop task. Table~\ref{tab:sim_unseen} reports the results. KAI substantially outperforms Seer, consistent with the interpretation that explicit kinematic reasoning facilitates generalization across geometric variations.


In the real world, we train on a single laptop instance and test on a different laptop instance. Table~\ref{tab:real_unseen} reports the results. KAI again outperforms DP3 and Seer, and the addition of video co-training provides a further gain, suggesting that human interaction data contributes complementary visual-kinematic experience that aids cross-instance transfer.

\end{document}